\documentclass{bluecardreport}

\usepackage{amsmath,amssymb,bm}
\usepackage{multirow}
\usepackage{makecell}
\usepackage{colortbl}
\usepackage{tabularx}
\usepackage{placeins}
\usepackage{adjustbox}
\usepackage{tikz}
\usepackage{wrapfig}
\usepackage{graphicx}
\usepackage{subcaption}
\usepackage{etoc}
\usetikzlibrary{arrows.meta,fit,positioning,shapes.geometric}

\definecolor{benchmarkgroupbg}{RGB}{240,244,249}
\reportname{Understanding Affective Adaptation in Multimodal Foundation Models: Emergent Functional Specialization}
\reportbrand{}
\reportlogo{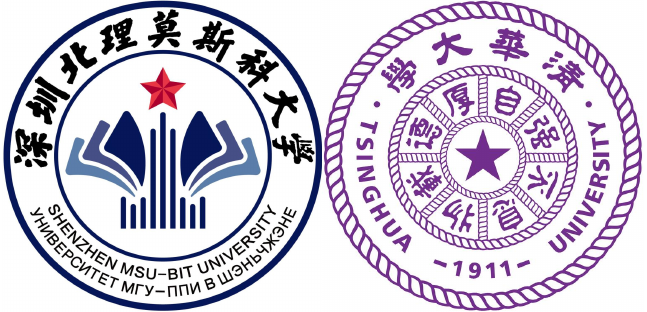}

\date{Technical Report --- September 29, 2026}

\title{Understanding Affective Adaptation in Multimodal Foundation Models:
Emergent Functional Specialization}
\reportauthor{Zhen Zhang$^{1}$, Runhao Zeng$^{1}$, Sicheng Zhao$^2$, Xiping Hu$^{1}$}

\reportaffiliation{$^1$Shenzhen MSU-BIT University,  
$^2$Tsinghua University}

\reportabstract{Despite rapid progress in multimodal affective foundation models, how affective capabilities emerge within their internal architectures remains poorly understood. A critical open question is whether affective fine-tuning induces diffuse changes across the model or organizes computation into functionally specialized pathways. 
We systematically investigate this question through a broad module-level
analysis of 13 affective model instances spanning nine model designs, multiple
scales, tasks, and training paradigms, complemented by controlled functional
analyses on representative models.
We find that affective adaptation exhibits a consistent yet non-exclusive functional organization. Under matched trainable-parameter budgets, adapting only the feed-forward network (FFN) consistently outperforms adapting only the attention modules across all evaluated settings and, on average, nearly matches the performance obtained by tuning all major Transformer projections, identifying the FFN as a particularly efficient adaptation substrate. 
More strikingly, although the gate, up, and down projections exhibit
comparable standalone adaptation capacity, their learned functional
contributions become differentiated after joint optimization. Module recovery
and targeted interventions identify \texttt{gate\_proj} as a particularly
prominent pathway, while checkpoint analysis shows that this differentiation
develops over the course of training. We characterize this phenomenon as
emergent functional specialization: distinct pathway-level roles are not
fully explained by standalone adaptation capacity, but arise through joint
affective adaptation.
Building on this finding, Gate-Focused Efficient Tuning (GET) retains 96.2-98.0\% of the performance obtained by tuning all major Transformer projections while using only 19.3-24.5\% as many trainable parameters.}
\reportteaser{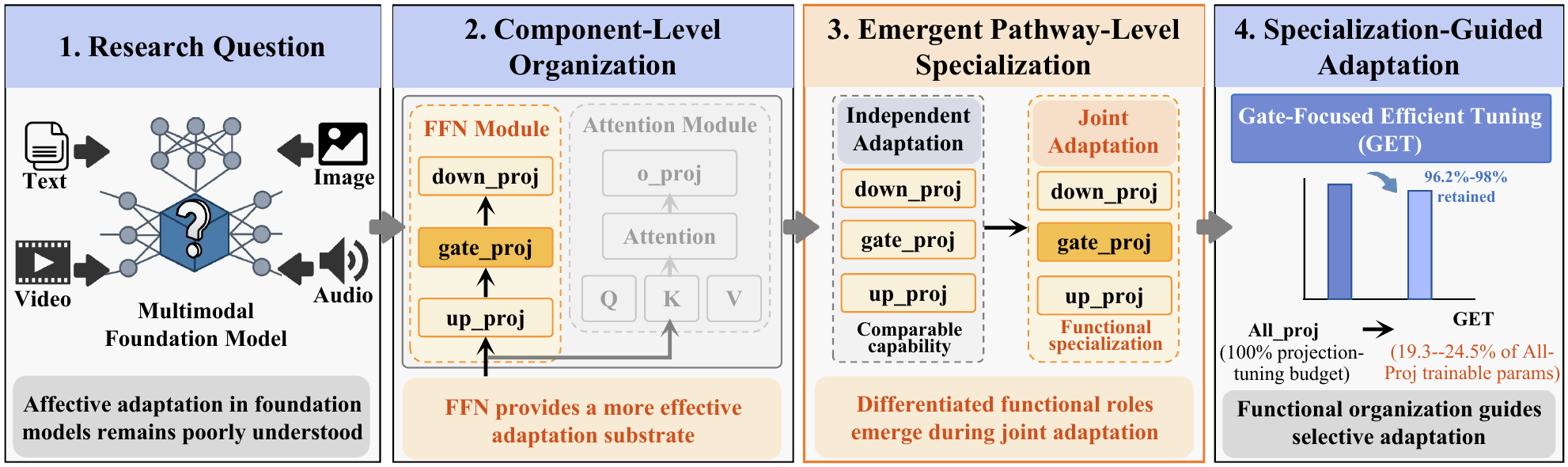}
\reportteasercaption[fig:affectgate-overview]{%
Overview of our study.
  We characterize affective adaptation from component-level organization to
  pathway-level functional specialization, examine how this specialization
  develops during joint optimization, and evaluate its practical utility for
  selective adaptation.
}

\begin{document}

\maketitle

\clearpage
\tableofcontents

\clearpage
\reportpagestyle

\section{Introduction}
\label{section:1}

% \begin{figure*}[!t]
%   \centering
%   \includegraphics[width=0.95\linewidth]{figures/overall.pdf}
%   % \vspace{-0.6cm}
%   \caption{
%   Overview of our study.
%   We characterize affective adaptation from component-level organization to
%   pathway-level functional specialization, examine how this specialization
%   develops during joint optimization, and evaluate its practical utility for
%   selective adaptation.
%   }
%   \label{fig:GET_step}
%   \vspace{-0.45cm}
% \end{figure*}

Large language models and multimodal foundation models have demonstrated
strong capabilities in affective understanding, including emotion recognition,
sentiment analysis, affective conversation, and depression detection.
Recent affective foundation models such as Emotion-LLaMA~\citep{cheng2024emotion},
AffectGPT~\citep{lian2025affectgpt}, OneEmo~\citep{huang2026oneemo}, and EAI-LLM~\citep{lu2025understanding}
further show that general-purpose foundation models can be effectively adapted
to a wide range of affect-related tasks across modalities.
Despite this progress, considerably less is known about what happens
\emph{inside the model during affective adaptation}. In particular, it remains
unclear how the acquired functionality is distributed across Transformer
components and whether different computational pathways assume distinct roles
through the adaptation process.

Recent mechanistic studies have shown that model capabilities can be associated
with structured internal representations, including specific neurons,
attention heads, pathways, and cross-layer circuits
~\citep{shao2025reasons,sofroniew2026emotion,wang2025llms}.
Such studies provide important evidence that learned capabilities need not be
uniformly distributed throughout a model. However, identifying structures
associated with an already acquired capability is different from understanding
\emph{how functional organization develops as that capability is acquired}.
For affective models, this process remains largely unexplored.
We therefore ask:
\textbf{how does affective adaptation become functionally organized across
Transformer pathways, and does this organization emerge through optimization?}

To answer this question, we conduct a broad analysis across 13 affective model
instances spanning nine model designs, multiple backbone families, model
scales, affective tasks, and adaptation paradigms. We use this broad model
collection to characterize module-level adaptation patterns, and then conduct
controlled pathway-level analyses on representative models to examine
functional specialization more closely.
We first examine adaptation at the level of major Transformer components.
Parameter changes are strongly non-uniform, with FFN projections generally
exhibiting larger adaptation-induced changes than Attention projections.
Under comparable trainable-parameter budgets, adapting the FFN also tends to achieve stronger affective performance than adapting Attention. These observations suggest that, in the models studied here, the FFN may serve as a more effective substrate for affective adaptation.

We then move from component-level organization to individual FFN pathways. Although \texttt{gate\_proj}, \texttt{up\_proj}, and \texttt{down\_proj} show broadly comparable standalone adaptation capacity, their functional contributions appear to diverge after joint optimization. Module recovery and targeted intervention results suggest that \texttt{gate\_proj} becomes a particularly prominent pathway, while checkpoint analysis indicates that this differentiation emerges over the course of training. Taken together, these observations suggest that the resulting specialization cannot be fully explained by inherent differences in standalone adaptation capacity; instead, distinct functional roles may emerge through joint affective adaptation.

Finally, we examine whether the functional organization identified above can provide practical guidance for selective adaptation. As a simple mechanism-inspired validation, we restrict LoRA adaptation to the prominent gating pathway, leading to Gate-Focused Efficient Tuning (GET). Across AffectGPT, Emotion-LLaMA, and EAI-LLM, GET retains \textbf{96.2-98.0\%} of the affective performance achieved by All-Proj while using only \textbf{19.3-24.5\%} of its trainable parameters. Comparable performance retention is not observed in the examined non-affective controls. These results suggest that the functional organization revealed by our analysis may offer useful guidance for selective and parameter-efficient affective adaptation. Our main contributions are summarized as follows:

(1) We provide a systematic pathway-level analysis of affective adaptation in
multimodal foundation models and show that it exhibits structured,
non-uniform organization. Under comparable trainable-parameter budgets,
the FFN supports stronger affective adaptation than Attention.

(2) 
We distinguish \emph{standalone adaptation capacity} from
\emph{functional role under joint optimization}. Although multiple FFN
pathways exhibit comparable standalone adaptation capacity, their functional
contributions appear to diverge under joint optimization, with this
differentiation emerging progressively over the course of training.

(3) 
We translate the observed specialization into
Gate-Focused Efficient Tuning (GET), which retains
96.2--98.0\% of All-Proj affective performance using only
19.3--24.5\% of its trainable-parameter budget across the examined affective
models.
\section{Related Work}
\label{section:2}

\paragraph{Affective Foundation Models}
Large language models and multimodal foundation models have been increasingly
adopted for affective computing tasks, including emotion recognition, sentiment
analysis, affective dialogue, and depression detection. Representative models
such as Emotion-LLaMA~\citep{cheng2024emotion},
AffectGPT~\citep{lian2025affectgpt}, and
EAI-LLM~\citep{lu2025understanding} demonstrate that large-scale foundation
models can support affective understanding across language, speech, image, and
video modalities. Recent studies further explore affective modeling under
different supervision, domain, and multimodal settings
~\citep{huang2025emodetective,chen2026towards,song2025emotion,yang2025echoes}.
Most existing work focuses on improving prediction accuracy, generation,
robustness, or multimodal integration. In comparison, considerably less is
known about how affective capabilities become functionally organized within
the Transformer backbone during adaptation. Our work focuses on this internal
organization rather than introducing a new affective task or model architecture.

\paragraph{Parameter-Efficient and Selective Adaptation}
Foundation models are commonly adapted to downstream tasks through
parameter-efficient fine-tuning (PEFT). LoRA
~\citep{hu2021loralowrankadaptationlarge}, for example, represents task-specific
updates using trainable low-rank factors while keeping pretrained parameters
frozen. Subsequent work explores non-uniform allocation of adaptation capacity.
AdaLoRA~\citep{zhang2023adalora} dynamically allocates rank budgets across
weight matrices according to their estimated importance, while other studies
investigate selective tuning over parameter subsets or Transformer layers
~\citep{ouyang2025k,zhou2025lora,wei2025flexora}.
These methods show that model components need not contribute equally to
downstream adaptation. However, most selective PEFT approaches allocate
capacity based on parameter importance, gradients, adapter outputs, layer
selection, or downstream performance. Our work instead characterizes the
capability-specific functional organization emerging during affective
adaptation and examines whether it can guide target-module selection.

\paragraph{Mechanistic Analysis of Model Capabilities}
Mechanistic studies have increasingly revealed structured internal organization
underlying learned capabilities
~\citep{shao2025reasons,sofroniew2026emotion,wang2025llms}.
For example, \citet{shao2025reasons} associate mathematical
reasoning with particular Transformer pathways, including the attention output
projection. In the affective domain, \citet{sofroniew2026emotion} study internal representations of emotion
concepts, while \citet{wang2025llms} identify emotion-related
neurons, attention heads, and cross-layer circuits associated with emotional
behavior. These studies mainly characterize structures associated with
already acquired capabilities. In contrast, we study how differentiated
functional roles develop across standard Transformer pathways during affective
adaptation, distinguishing a pathway's standalone adaptation capacity from the
role it assumes under joint optimization.
\section{How Is Affective Adaptation Organized across Transformer Components?}
\label{section:3}

\subsection{Analyzed Transformer Components}

We focus on the Transformer language backbones of the affective foundation
models examined in this work. These models are typically adapted from
general-purpose LLMs using parameter-efficient techniques such as LoRA,
where the major linear projections in Attention and the FFN constitute
common adaptation targets. Moreover, the Qwen- and LLaMA-based backbones
examined in our experiments share a common projection structure, providing
a consistent architectural basis for cross-model comparison.

\begin{wrapfigure}{r}{0.46\columnwidth}
    \centering
    \vspace{-2.0em}
    \includegraphics[width=\linewidth]{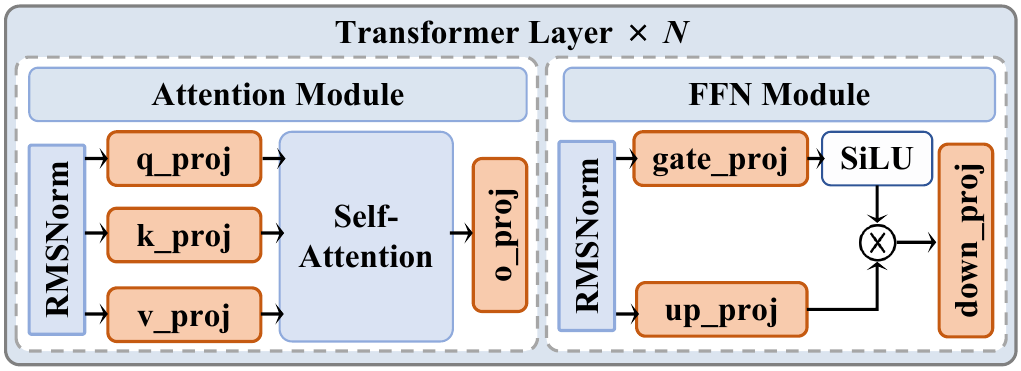}
    % \vspace{-0.6cm}
    \caption{
    Schematic of the Transformer block analyzed in this work.
    Colored boxes denote the projection pathways considered in our
    module-level analysis.
    }
    \label{fig:transformer_architecture}
\end{wrapfigure}

As illustrated in Figure~\ref{fig:transformer_architecture}, we analyze
seven projection pathways: \texttt{q\_proj}, \texttt{k\_proj},
\texttt{v\_proj}, and \texttt{o\_proj} in Attention, and
\texttt{gate\_proj}, \texttt{up\_proj}, and \texttt{down\_proj}
in the gated FFN. These pathways allow us to compare adaptation both
between the two major Transformer components (Attention versus FFN)
and among their constituent projections.

Our analysis focuses on the shared Transformer backbone rather than
attempting to characterize every modality-specific encoder or projector
in the multimodal architecture.

\subsection{Affective Adaptation Produces Non-Uniform Parameter Changes}

We first visualize four representative models---AffectGPT~\citep{lian2025affectgpt},
Emotion-LLaMA~\citep{cheng2024emotion}, ESCRAG-R1~\citep{liu2026escrag}, and EAI-LLM~\citep{lu2025understanding}---covering different backbone families,
affective tasks, and adaptation settings. The broader analysis across 13 model
instances is provided in Appendix~\ref{appendix:A}.
For each projection $X$, we quantify the adaptation-induced parameter change
using the Frobenius norm:
$
\Delta_X =
\left\|
W_X^{(\mathrm{Affect})}
-
W_X^{(\mathrm{Base})}
\right\|_F
$,
where $W_X^{(\mathrm{Affect})}$ and $W_X^{(\mathrm{Base})}$ denote the
weights of projection $X$ in the adapted and base models, respectively.
We compute $\Delta_X$ at each Transformer layer to characterize how
parameter changes vary across projection pathways and network depth. As the
raw Frobenius norm may also reflect differences in projection dimensionality,
we treat it as a measure of overall weight change rather than per-parameter
update magnitude.

\begin{figure*}[t]
  \centering

  \begin{minipage}[t]{0.49\linewidth}
    \centering
    \includegraphics[width=\linewidth]{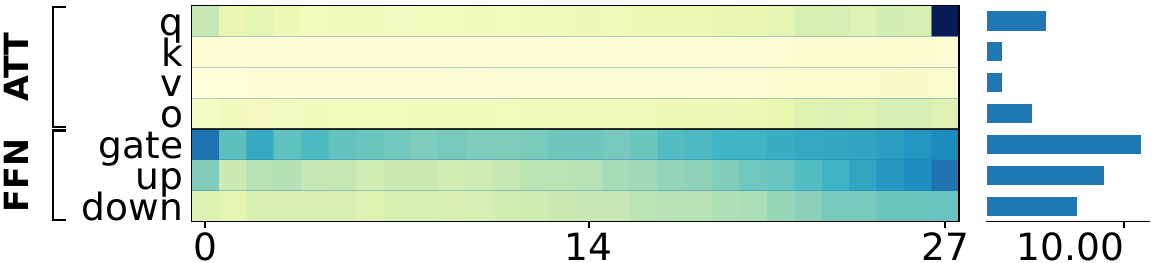}
    \vspace{0.25em}
    {\small (a) Qwen2.5-7B-Instruct $\xrightarrow{\mathrm{adapt}}$ AffectGPT}
  \end{minipage}
  \hfill
  \begin{minipage}[t]{0.49\linewidth}
    \centering
    \includegraphics[width=\linewidth]{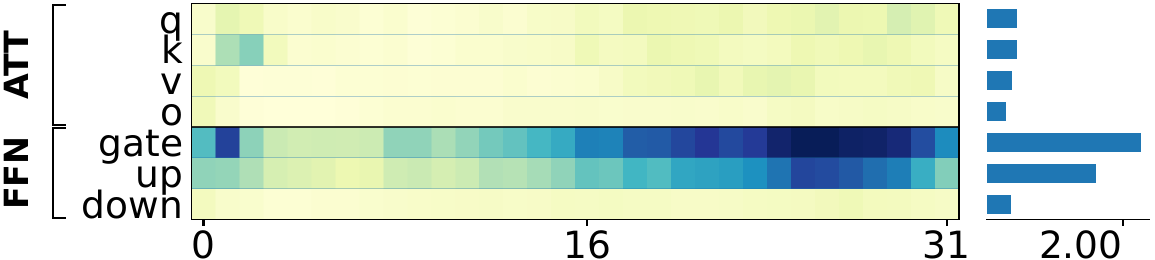}
    \vspace{0.25em}
    {\small (b) LLaMA-2-7B-chat-hf $\xrightarrow{\mathrm{adapt}}$ Emotion-LLaMA}
  \end{minipage}

  \vspace{0.8em}

  \begin{minipage}[t]{0.49\linewidth}
    \centering
    \includegraphics[width=\linewidth]{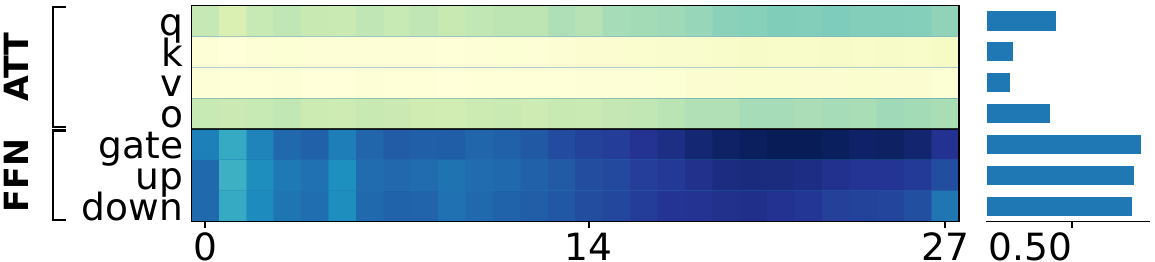}
    \vspace{0.25em}
    {\small (c) Qwen2.5-7B-Instruct $\xrightarrow{\mathrm{adapt}}$ ESCRAG-R1}
  \end{minipage}
  \hfill
  \begin{minipage}[t]{0.49\linewidth}
    \centering
    \includegraphics[width=\linewidth]{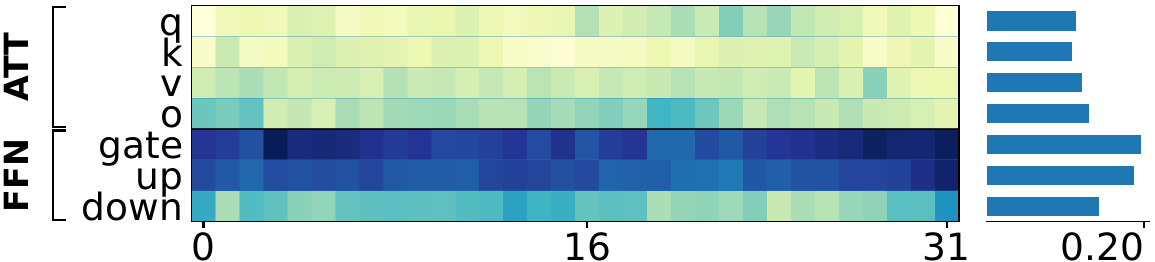}
    \vspace{0.25em}
    {\small (d) LLaMA-2-7B-chat-hf $\xrightarrow{\mathrm{adapt}}$ EAI-LLM}
  \end{minipage}

  \caption{
  Module-wise parameter changes induced by affective adaptation across four representative models. Heatmaps show layer-wise Frobenius norms of weight differences, while bars report layer-averaged values.
  Across the examined models, the updates are non-uniformly distributed, with
  FFN projections generally exhibiting larger changes than Attention projections.
  }
  \label{fig:3.1}
\end{figure*}

As shown in Figure~\ref{fig:3.1}, the resulting changes are clearly
non-uniform across Transformer components. Despite differences in backbone
architecture and network depth, all four representative models exhibit a
similar coarse-grained pattern: FFN projections generally undergo larger
parameter changes than Attention projections. This consistency suggests that
affective adaptation modifies different Transformer components to different
extents rather than perturbing them uniformly.

We further assess robustness using complementary element-wise statistics and extend the analysis across additional model scales, architectures, tasks, and adaptation paradigms (Appendix~\ref{appendix:A}--\ref{appendix:Non-Affective Adaptation}). The same coarse-grained asymmetry remains observable, suggesting that the pattern is not specific to the global Frobenius metric or the four representative models considered here. Importantly, however, larger parameter changes do not necessarily indicate greater functional relevance.

\subsection{The FFN Provides a More Effective Adaptation Substrate}

The parameter-space analysis above reveals systematically larger changes in
the FFN, but update magnitude alone does not establish whether the FFN is
functionally more effective for affective adaptation. We therefore compare
Attention and FFN under controlled trainable-parameter budgets.

Specifically, we conduct matched-budget adaptation experiments on AffectGPT,
Emotion-LLaMA, and EAI-LLM under three settings: Attention-only (ATT-Only),
FFN-only (FFN-Only), and all projection modules (All-Proj). ATT-Only and
FFN-Only use comparable numbers of trainable parameters, enabling a controlled
comparison under similar adaptation budgets. Since the models employ different
evaluation metrics, the Average row is reported only as an unweighted
descriptive macro-average and should not be interpreted as a unified metric.

As shown in Table~\ref{tab:att_ffn_comparison}, FFN-Only consistently
outperforms ATT-Only across all three models, yielding an average improvement
of 1.33 points under comparable trainable-parameter budgets.
At the same time, FFN-Only retains 99.2\% of the All-Proj performance on
average and even slightly exceeds All-Proj on EAI-LLM.
Taken together, these results show that, under comparable trainable-parameter
budgets, the FFN supports stronger affective adaptation than Attention while
approaching the performance of All-Proj adaptation.

\begin{table}[t]
\centering
\caption{
Performance of Attention-only and FFN-only adaptation under comparable trainable-parameter budgets, with All-Proj as the reference. Parentheses indicate trainable parameter counts and their percentages of the full model.
}
% \vspace{-0.3cm}
\label{tab:att_ffn_comparison}
\resizebox{\linewidth}{!}{
\begin{tabular}{lccccc}
\toprule
\textbf{Model} &
\textbf{ATT-only} &
\textbf{FFN-only} &
$\boldsymbol{\Delta}_{\mathrm{FFN-ATT}}$ &
\textbf{All-Proj} &
\textbf{FFN / All}
\\
\midrule

AffectGPT &
73.91 \
\newline
(10.09M, 0.13\%) &
\textbf{74.83} \
\newline
(9.46M, 0.12\%) &
\textbf{+0.92} &
76.30 &
98.1\%
\\

Emotion-LLaMA &
86.60 \
\newline
(93.32M, 1.37\%) &
\textbf{87.71} (92.79M, 1.36\%) &
\textbf{+1.11} &
88.81 &
98.8\%
\\

EAI-LLM &
71.78 \
\newline
(93.32M, 1.37\%) &
\textbf{73.73} \
\newline
(92.79M, 1.36\%) &
\textbf{+1.95} &
73.01 &
101.0\%
\\

\midrule

\textbf{Average} &
77.43 &
\textbf{78.76} &
\textbf{+1.33} &
79.37 &
99.2\%
\\

\bottomrule
\end{tabular}
}
\end{table}
\section{How Does Functional Specialization Arise Within the FFN?}
\label{section:4}

Having established a component-level advantage of the FFN, we next examine
how its internal pathways acquire distinct functional roles.
We first compare their standalone adaptation capacity and then analyze how
their contributions change under joint optimization.

\subsection{Multiple FFN Pathways Exhibit Comparable Standalone Adaptation Capacity}
\label{ffn-train}

We first ask whether functional specialization within the FFN simply reflects
an intrinsic difference in standalone adaptation capacity. To examine this
question, we independently optimize each projection while keeping all other
model parameters frozen. This includes the four Attention projections
(\texttt{q\_proj}, \texttt{k\_proj}, \texttt{v\_proj}, and \texttt{o\_proj})
and the three FFN projections (\texttt{gate\_proj}, \texttt{up\_proj}, and
\texttt{down\_proj}). The full set of projections provides a common reference
across the Transformer, while our main focus is whether the three FFN
pathways differ substantially in their standalone affective adaptation
capacity.

\begin{figure}[t]
% \vspace{-1.5cm}
  \centering
  \begin{subfigure}[t]{0.48\linewidth}
    \centering
    \includegraphics[width=\linewidth]{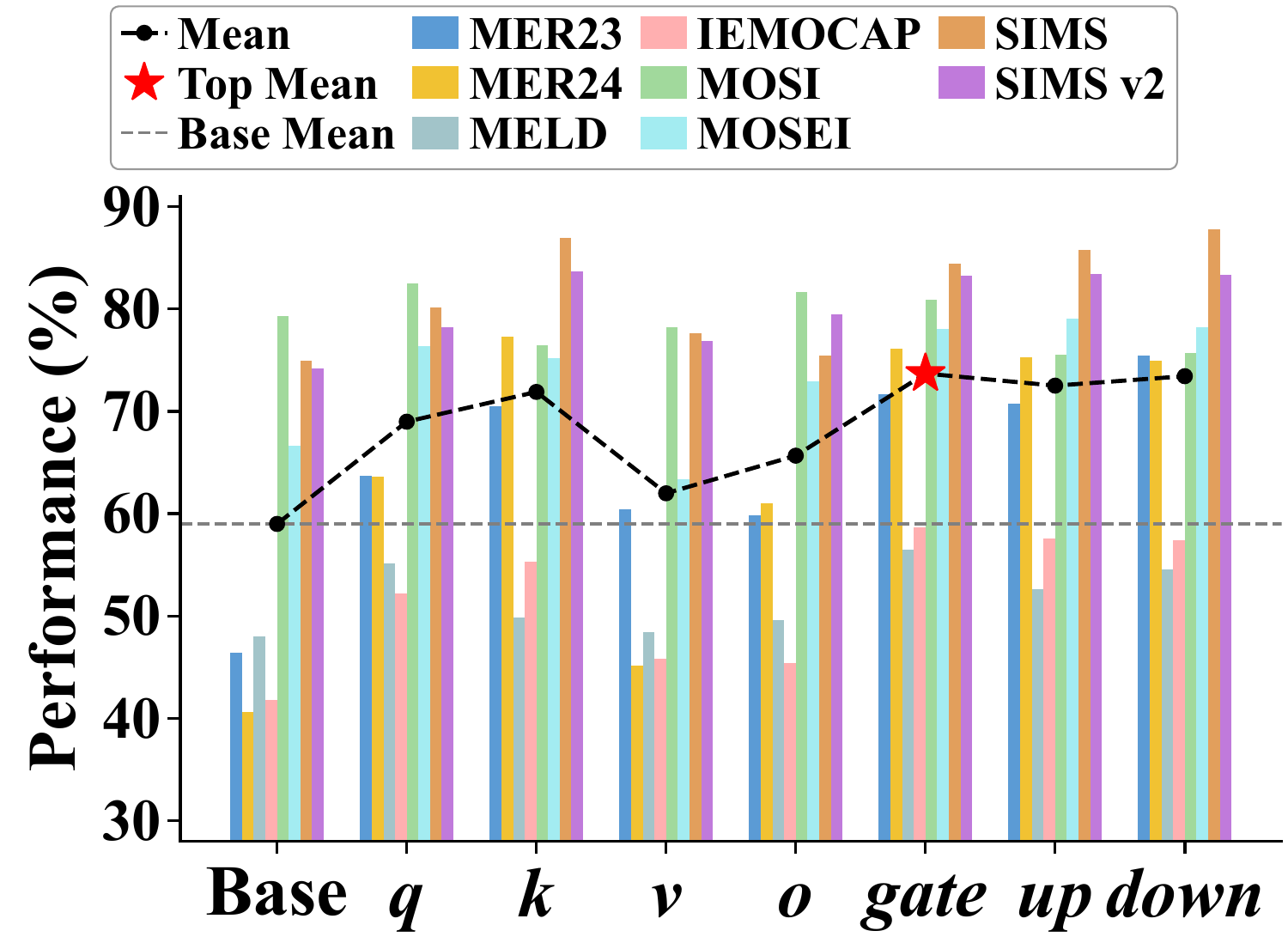}
    \caption{Standalone adaptation capacity}
    \label{fig:diff_modules_train}
  \end{subfigure}%
  \hfill
  \begin{subfigure}[t]{0.48\linewidth}
    \centering
    \includegraphics[width=\linewidth]{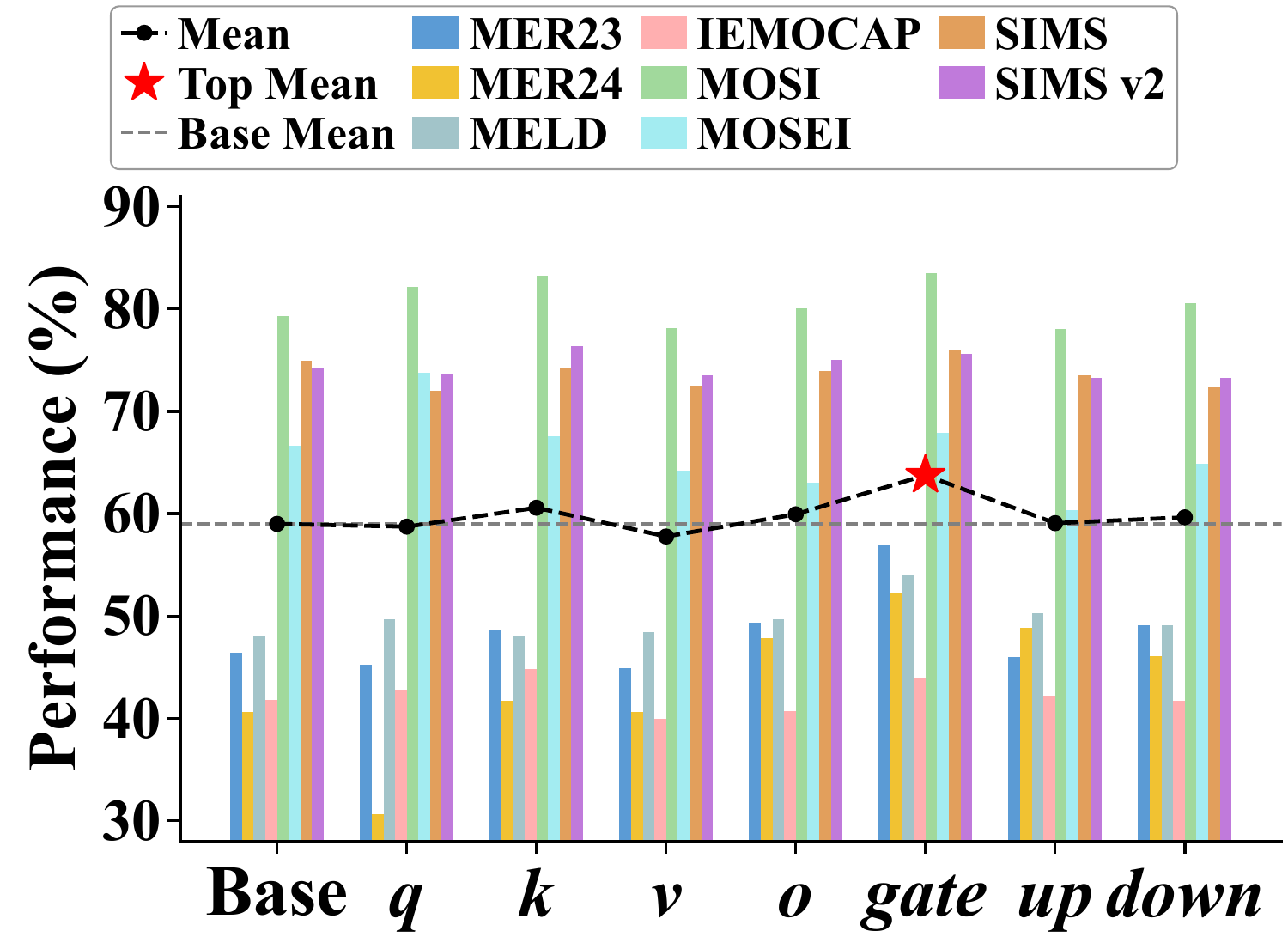}
    \caption{Learned functional contribution}
    \label{fig:diff_modules_lora}
  \end{subfigure}
\vspace{-0.2cm}
\caption{
Standalone adaptation capacity and learned functional contribution across Transformer projections. (a) FFN pathways achieve comparable performance when optimized independently. (b) After joint adaptation, their functional roles differentiate, with \texttt{gate\_proj} achieving the highest mean recovery performance.
}
  \label{fig:functional_specialization}
\end{figure}

As shown in Figure~\ref{fig:diff_modules_train}, all three FFN pathways
achieve strong affective performance when optimized independently and
outperform most Attention projections. More importantly, their average
performance remains relatively close. Although minor differences exist, no single FFN pathway exhibits an
overwhelming advantage under independent adaptation, indicating broadly
comparable standalone adaptation capacity among the three pathways.

\subsection{Jointly Adapted Pathways Exhibit Differentiated Functional Contributions}

We next examine the functional roles assumed by individual pathways after
joint adaptation. Starting from the fully adapted All-Proj model, we separately
recover the learned LoRA updates of each projection in the corresponding base
model, while leaving all other adaptation updates unloaded. The resulting
performance measures how much of the jointly learned affective functionality
can be recovered from each pathway in isolation.

As shown in Figure~\ref{fig:diff_modules_lora}, the pattern differs markedly
from standalone adaptation. While the three FFN pathways exhibit broadly
comparable adaptation capacity when optimized independently, their learned
contributions become clearly differentiated after joint optimization.
In particular, recovering only the \texttt{gate\_proj} updates yields the
highest mean performance, whereas \texttt{up\_proj} and
\texttt{down\_proj} recover less.

This contrast shows that comparable standalone adaptation capacity does not
necessarily translate into comparable functional roles after joint
optimization. Instead, the jointly adapted model develops an asymmetric
distribution of learned functionality across FFN pathways, with
\texttt{gate\_proj} emerging as the most prominent pathway in this recovery
analysis.

\subsection{Targeted Interventions Reveal Uneven Functional Dependency}
\label{sec:intervention}

The recovery analysis above characterizes how much jointly learned
functionality can be recovered from each pathway in isolation, but it does not
directly measure how strongly the fully adapted model depends on that pathway.
We therefore complement the recovery analysis with targeted interventions on
the learned projection-wise LoRA updates.

For each projection pathway, we apply three interventions to its learned update
$\Delta W$.
\textbf{Removal} sets the corresponding update to zero;
\textbf{Randomization} replaces the learned update with randomized values; and
\textbf{Direction Reversal} replaces $\Delta W$ with $-\Delta W$.
All other learned updates remain unchanged.
We then measure the resulting degradation in affective performance relative
to the original All-Proj model.

\begin{table*}[t]
    \centering
\caption{
Targeted interventions on projection-wise LoRA updates in AffectGPT and OneEmo.
Values in parentheses denote absolute performance changes relative to the All-Proj baseline.
}
% \vspace{-0.3cm}
    \label{tab:module_intervention_results}
    \small
    \setlength{\tabcolsep}{8pt}
    \renewcommand{\arraystretch}{1.15}

    \begin{tabular}{llccc}
        \toprule
        \textbf{Model} &
        \textbf{Module} &
        \textbf{Removal} &
        \textbf{Randomization} &
        \textbf{Direction Reversal} \\
        \midrule

        \multirow{8}{*}{AffectGPT}
        & \texttt{All-Proj}
        & 76.30
        & 76.30
        & 76.30 \\
        \cmidrule(lr){2-5}

        & \texttt{q\_proj}
        & 68.16 ($\downarrow$8.14)
        & 67.67 ($\downarrow$8.63)
        & 65.20 ($\downarrow$11.10) \\

        & \texttt{k\_proj}
        & 68.78 ($\downarrow$7.52)
        & 68.46 ($\downarrow$7.84)
        & 69.39 ($\downarrow$6.91) \\

        & \texttt{v\_proj}
        & 69.07 ($\downarrow$7.23)
        & 69.25 ($\downarrow$7.05)
        & 69.05 ($\downarrow$7.25) \\

        & \texttt{o\_proj}
        & 69.69 ($\downarrow$6.61)
        & 68.94 ($\downarrow$7.36)
        & 69.63 ($\downarrow$6.67) \\

        & \textbf{\texttt{gate\_proj}}
        & \textbf{65.66}
          \textbf{($\downarrow$10.64)}
        & \textbf{66.20}
          \textbf{($\downarrow$10.10)}
        & \textbf{55.97}
          \textbf{($\downarrow$20.33)} \\

        & \texttt{up\_proj}
        & 68.99 ($\downarrow$7.31)
        & 68.88 ($\downarrow$7.42)
        & 67.03 ($\downarrow$9.27) \\

        & \texttt{down\_proj}
        & 69.04 ($\downarrow$7.26)
        & 69.63 ($\downarrow$6.67)
        & 68.35 ($\downarrow$7.95) \\

\midrule

        \multirow{8}{*}{OneEmo}
        & \texttt{All-Proj}
        & 71.80
        & 71.80
        & 71.80 \\
        \cmidrule(lr){2-5}

        & \texttt{q\_proj}
        & 69.95 ($\downarrow$1.85)
        & 70.12 ($\downarrow$1.68)
        & 68.15 ($\downarrow$3.65) \\

        & \texttt{k\_proj}
        & 71.70 ($\downarrow$0.10)
        & 72.07 ($\uparrow$0.27)
        & 72.00 ($\uparrow$0.20) \\

        & \texttt{v\_proj}
        & 72.48 ($\uparrow$0.68)
        & 71.54 ($\downarrow$0.26)
        & 70.87 ($\downarrow$0.93) \\

        & \texttt{o\_proj}
        & 71.00 ($\downarrow$0.80)
        & 71.43 ($\downarrow$0.37)
        & 71.88 ($\uparrow$0.08) \\

        & \textbf{\texttt{gate\_proj}}
        & \textbf{66.33 ($\downarrow$5.47)}
        & \textbf{66.14 ($\downarrow$5.66)}
        & \textbf{62.80 ($\downarrow$9.00)} \\

        & \texttt{up\_proj}
        & 71.62 ($\downarrow$0.18)
        & 71.47 ($\downarrow$0.33)
        & 68.50 ($\downarrow$3.30) \\

        & \texttt{down\_proj}
        & 71.31 ($\downarrow$0.49)
        & 70.86 ($\downarrow$0.94)
        & 70.00 ($\downarrow$1.80) \\

        \bottomrule
    \end{tabular}
\end{table*}

As shown in Table~\ref{tab:module_intervention_results}, intervention
sensitivity varies substantially across projection pathways, with a consistent
pattern across both models. In AffectGPT, disrupting \texttt{gate\_proj}
produces the largest performance degradation under all three interventions.
The effect is particularly pronounced under Direction Reversal, where replacing
its learned update $\Delta W$ with $-\Delta W$ reduces performance from 76.30
to 55.97, a drop of 20.33 points.

A similar pattern emerges in OneEmo. Removing or randomizing the
\texttt{gate\_proj} update reduces performance by 5.47 and 5.66 points,
respectively, substantially more than interventions on the other projections.
Direction Reversal again results in the largest performance degradation for \texttt{gate\_proj}, causing a 9.00-point drop. In contrast, interventions on
several Attention projections have relatively small effects and occasionally
yield slight performance improvements. These results indicate that the stronger
dependency on the gating pathway is not specific to AffectGPT.

The intervention results provide evidence complementary to the recovery
analysis. Whereas recovery measures how much jointly learned functionality can
be recovered from a pathway in isolation, intervention measures how sensitive
the fully adapted model is to disrupting that pathway. Together, these analyses
provide converging evidence that joint affective adaptation produces
differentiated functional roles across otherwise capable FFN pathways, with
\texttt{gate\_proj} emerging as a particularly prominent contributor.

\subsection{Functional Differentiation Develops During Affective Adaptation}
\label{sec:functional_dynamics}

\begin{figure}[!t]
  \centering
  \begin{subfigure}[t]{0.48\linewidth}
    \centering
    \includegraphics[width=\linewidth]{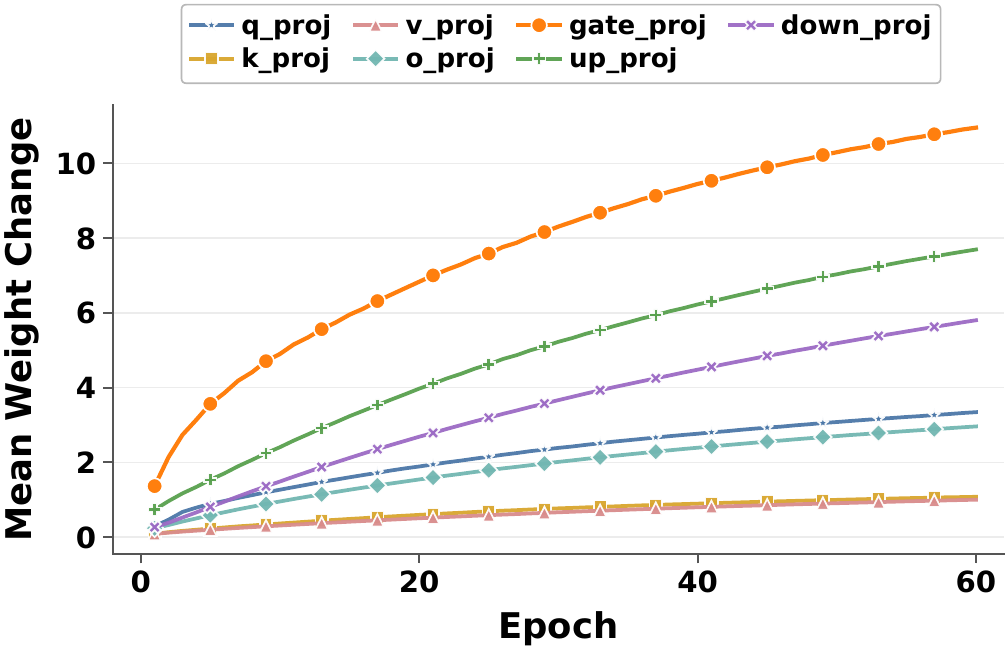}
    \caption{Evolution of module-wise LoRA updates}
    \label{fig:module_update_dynamics}
  \end{subfigure}%
  \hfill
  \begin{subfigure}[t]{0.48\linewidth}
    \centering
    \includegraphics[width=\linewidth]{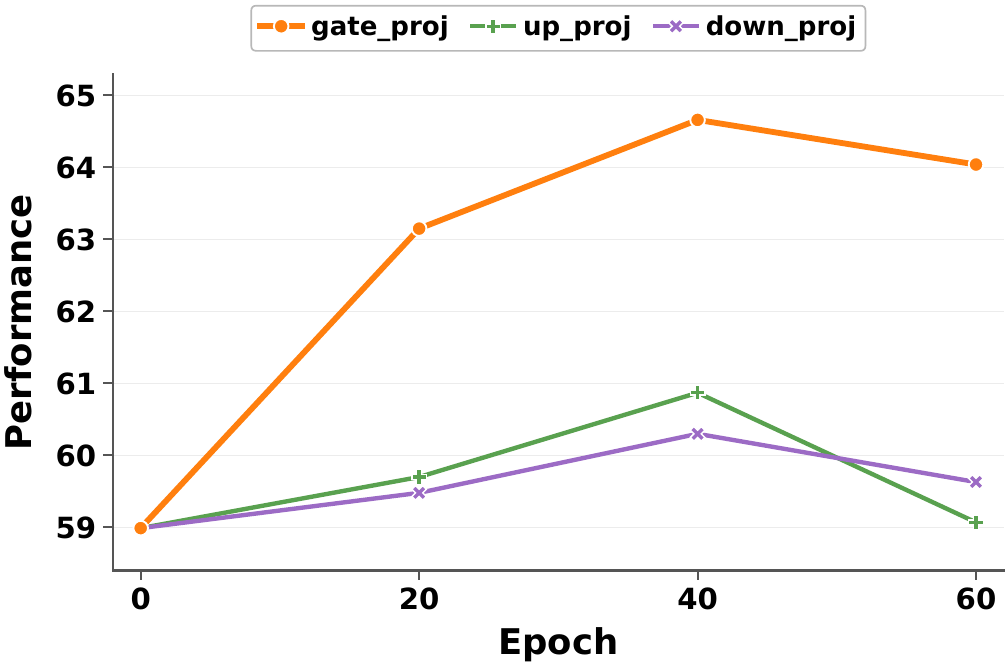}
    \caption{Epoch-wise recovery of FFN pathways}
    \label{fig:ffn_recovery_dynamics}
  \end{subfigure}

  \caption{
  Training dynamics of pathway differentiation during affective adaptation.
  (a) LoRA update magnitudes across Transformer projections become increasingly
  differentiated over training.
  (b) Checkpoint-wise recovery of the three FFN pathways reveals emerging
  functional differentiation, with \texttt{gate\_proj} becoming more prominent
  during training.
  }
  \label{fig:functional_dynamics}
\end{figure}

The preceding analyses characterize differentiated functional roles among FFN
pathways in the final adapted model. We next examine how these differences
develop during joint optimization by tracking parameter changes across
Transformer projections and functional recovery of the three FFN pathways
throughout training.

We first examine the evolution of pathway-wise parameter updates. As shown in
Figure~\ref{fig:module_update_dynamics}, the learned LoRA update magnitudes
become increasingly differentiated over training. In this trajectory,
\texttt{gate\_proj} accumulates the largest changes, while the other
projections follow distinct optimization trajectories. A similar
differentiation of parameter updates is observed in an additional model
(Appendix~\ref{appendix:training_dynamics}), although the relative ordering
among FFN pathways varies across models.

We then ask whether this parameter-level differentiation is accompanied by
changes in functional contribution. For \texttt{gate\_proj},
\texttt{up\_proj}, and \texttt{down\_proj}, we recover their learned LoRA
updates from intermediate checkpoints and evaluate each pathway in isolation.
As shown in Figure~\ref{fig:ffn_recovery_dynamics}, the three pathways exhibit
similar recovery performance at early stages, but their contributions diverge
as optimization proceeds. By intermediate epochs, \texttt{gate\_proj}
recovers noticeably higher performance than \texttt{up\_proj} and
\texttt{down\_proj}. Although the gap narrows toward the final checkpoint,
\texttt{gate\_proj} remains the strongest recovered FFN pathway.

These trajectories show that the functional differentiation observed
after joint adaptation is not apparent at early stages but develops during
optimization. Combined with the comparable standalone capacity observed in
Section~\ref{ffn-train}, this suggests the distinct roles of FFN pathways
are shaped through joint adaptation rather than simply inherited from
differences in their individual adaptation capacity.
\section{From Emergent Specialization to Efficient Affective Adaptation}
\label{section:5}

The preceding analyses reveal that joint affective adaptation develops
differentiated functional roles across FFN pathways, with
\texttt{gate\_proj} exhibiting a particularly prominent contribution.
We next examine whether this empirical specialization can provide useful
guidance for parameter-efficient affective adaptation.
Rather than treating the gating pathway as an inherently optimal or predefined
affective module, we use the observed functional organization as a basis for
selective adaptation.

\subsection{Gate-Focused Efficient Tuning}
\label{sec:get}

Based on this observation, we introduce
\textbf{Gate-Focused Efficient Tuning (GET)}, a simple selective adaptation
strategy that restricts trainable LoRA updates to the FFN gating pathway.
Specifically, GET attaches LoRA adapters only to \texttt{gate\_proj} in each
Transformer layer, while the Attention projections
(\texttt{q\_proj}, \texttt{k\_proj}, \texttt{v\_proj}, and
\texttt{o\_proj}) and the remaining FFN projections
(\texttt{up\_proj} and \texttt{down\_proj}) receive no trainable adapters.
During affective adaptation, only the LoRA parameters associated with
\texttt{gate\_proj} are optimized.

GET therefore serves as a direct test of whether the functional specialization
identified in the preceding analyses can be translated into a substantially
smaller adaptation budget while preserving affective performance.

\subsection{Performance Retention and Non-Affective Controls}
\label{sec:get_evaluation}

We evaluate whether the specialization-guided selection used by GET can
preserve affective performance under a substantially reduced trainable
parameter budget. For each model, GET is compared with All-Proj adaptation,
and we report both the fraction of trainable parameters used by GET and its
performance retention relative to the corresponding All-Proj baseline.

For the affective settings, we follow the original fine-tuning and evaluation
protocol of each model. AffectGPT is evaluated by averaging performance across
eight affective benchmarks, Emotion-LLaMA is evaluated on its original
benchmark, and EAI-LLM is evaluated by averaging performance across three
bodily emotion recognition datasets. Dataset descriptions and evaluation
metrics are provided in Appendix~\ref{appendix:C} and
Appendix~\ref{appendix:D}, respectively. Because the models use different
tasks and evaluation metrics, absolute scores are not compared across models;
instead, performance retention is computed within each model relative to its
own All-Proj baseline.

\begin{table*}[t]
\centering
\caption{
Performance retention and trainable-parameter efficiency of GET across affective and non-affective settings. Param. Ratio is relative to All-Proj, and Retention is the GET-to-All-Proj performance ratio.
}
\vspace{-0.3cm}
\label{tab:get_efficiency}

\small
\setlength{\tabcolsep}{3.5pt}
\renewcommand{\arraystretch}{1.15}

\begin{tabularx}{\textwidth}{
@{}
>{\raggedright\arraybackslash}m{0.14\textwidth}
>{\centering\arraybackslash}m{0.19\textwidth}
>{\centering\arraybackslash}m{0.22\textwidth}
>{\centering\arraybackslash}m{0.10\textwidth}
>{\centering\arraybackslash}m{0.10\textwidth}
>{\centering\arraybackslash}m{0.14\textwidth}
@{}
}
\toprule
\textbf{Setting} &
\textbf{Model} &
\centering\arraybackslash\textbf{Evaluation} &
\makecell{\textbf{Param.}\\\textbf{Ratio}} &
\textbf{Retention} &
\makecell{\textbf{GET /}\\\textbf{All-Proj}} \\
\midrule

\multirow{3}{*}{\makecell[l]{Affective\\Understanding}}
& AffectGPT
& \centering\arraybackslash 8 datasets (avg.)
& 24.5\%
& \textbf{96.6\%}
& 73.68 / 76.30 \\

& Emotion-LLaMA
& \centering\arraybackslash 1 dataset
& 19.3\%
& \textbf{98.0\%}
& 87.03 / 88.81 \\

& EAI-LLM
& \centering\arraybackslash 3 datasets (avg.)
& 19.3\%
& \textbf{96.2\%}
& 70.25 / 73.01 \\

\midrule

Mathematical Reasoning
& LLaMA (Gaussian)
& \centering\arraybackslash 2 datasets (avg.)
& 19.3\%
& 79.7\%
& 19.27 / 24.18 \\

\midrule

Instruction Following
& LLaMA (QLoRA)
& \centering\arraybackslash 1 dataset
& 19.3\%
& 74.0\%
& 32.93 / 44.50 \\

\bottomrule

\end{tabularx}
\end{table*}

As shown in Table~\ref{tab:get_efficiency}, GET retains 96.2--98.0\% of the
corresponding All-Proj performance across the three affective models while
using only 19.3--24.5\% of the All-Proj trainable-parameter budget.
This consistent within-model retention indicates that the functional
specialization identified above can provide a useful basis for substantially
reducing the adaptation budget without sacrificing most of the acquired
affective performance.

We additionally evaluate the same gate-focused strategy in two non-affective control settings, mathematical reasoning and instruction following, following the experimental protocols of~\citep{meng2024pissa,dettmers2023qlora}, respectively.
Under a comparable 19.3\% parameter ratio, GET retains only 79.7\% and 74.0\%
of the corresponding All-Proj performance, respectively. Thus, the high
retention observed in the affective models is not uniformly reproduced in
these non-affective controls. While these comparisons are not sufficient to
establish that gate-focused adaptation is uniquely specific to affective
capabilities, they provide a useful boundary test against the alternative
explanation that restricting adaptation to \texttt{gate\_proj} is generally
effective across downstream tasks.

The above results show that the pathway specialization uncovered
by our analysis has practical value: using \texttt{gate\_proj} as the sole
adaptation target preserves most of the All-Proj performance across the
examined affective models with a substantially smaller trainable-parameter
budget.

\section{Discussion}
\label{section:6}

\paragraph{Interpretation of Emergent Functional Specialization}
Our results highlight a distinction between a pathway's \emph{capacity} to
support affective adaptation and the \emph{functional role} it assumes under
joint optimization. Although the FFN pathways exhibit broadly comparable
standalone adaptation capacity, their contributions become differentiated
when optimized together. The prominence of \texttt{gate\_proj} should
therefore not be interpreted as evidence for a predefined ``emotion module,''
but rather as an emergent property of joint adaptation.

This perspective suggests that understanding post-training requires not only
asking which components can acquire a capability, but also how joint
optimization distributes learned functionality across interacting pathways.
The functional organization emerging after adaptation need not be predictable
from the standalone capacity of individual components.

The computational role of gating offers one possible interpretation.
Because gated FFNs multiplicatively modulate intermediate features, affective
adaptation may benefit from selectively emphasizing or suppressing existing
representations rather than uniformly rewriting them. This view is
conceptually consistent with accounts of affective modulation in neuroscience
~\citep{anderson2001lesions,janak2015circuits}. However, this remains a
hypothesis: our experiments do not establish selective modulation as the
underlying mechanism, nor imply a direct correspondence between Transformer
gating and biological affective processing.

\paragraph{Scope and Limitations}
Our analysis operates at the level of Transformer projection pathways and
does not yet explain affective processing at the activation, feature, or
circuit level. Moreover, the prominence of \texttt{gate\_proj} should not be
assumed to generalize universally across architectures, tasks, or post-training
procedures. Broader model families and finer-grained mechanistic analyses will
be needed to determine the generality and underlying mechanism of the observed
specialization.
\section{Conclusion}
\label{section:7}

This work investigates how affective adaptation becomes functionally organized
within multimodal foundation models. Across the examined settings, the FFN
supports stronger affective adaptation than Attention under comparable
trainable-parameter budgets, while its internal pathways exhibit comparable
standalone capacity but differentiated functional contributions after joint
optimization. Training dynamics further show that this differentiation
develops during adaptation, with \texttt{gate\_proj} emerging as a particularly
prominent pathway. Building on this organization, GET retains 96.2--98.0\% of
the corresponding All-Proj affective performance using only 19.3--24.5\% of
its trainable-parameter budget.

Overall, our findings suggest that affective capability is not simply localized
in a fixed architectural module, but can become \emph{functionally organized
through adaptation}. More broadly, we hope this perspective encourages a shift
from asking only \emph{where} a learned capability resides to understanding
\emph{how} its functional organization emerges during learning, providing a
starting point for future studies of capability formation, interpretation, and
selective adaptation in foundation models.

\subsection*{AI use statement}
Generative AI tools were used to assist with language compression and \LaTeX{}  formatting. The authors remain responsible for verifying all technical content, citations, claims, and reported results.

\subsection*{Ethics Statement}
All emotion-related datasets used in this work are publicly available research datasets and are used solely for academic research purposes. We do not collect new data from human participants or access private or personally identifiable information beyond what is included in the publicly released datasets. Our experiments focus on analyzing model adaptation and do not involve direct interactions with human subjects or deployment in real-world decision-making settings. We use all datasets in accordance with their intended research purposes and applicable licenses or terms of use. To the best of our knowledge, this work does not introduce additional ethical concerns beyond those associated with the use of existing public datasets. The authors declare no conflicts of interest related to this work.

\subsection*{Reproducibility statement}
To facilitate reproducibility, we provide our code in an anonymous repository at \href{https://anonymous.4open.science/r/AffectGate-98E4/README.md}{anonymous repository}. 
Sections~\ref{section:3}--\ref{section:5} examine the allocation and evolution of adaptation pathways in affective LLMs from complementary perspectives. The appendix presents additional experiments that complement the main findings, together with detailed descriptions of the datasets, models, and evaluation metrics used in our experiments.
% \newpage

\section{Appendix}
\label{sec:appendix}

\subsection{Module-Level Adaptation Patterns across Model Families and Scales}
\label{appendix:A}
To examine whether the observed module-level adaptation pattern generalizes across model families and scales, we extend our analysis to multiple model sizes within both the AffectGPT and Emotion-LLaMA series. Specifically, we analyze Qwen2.5-based AffectGPT models at 1.5B, 7B, 14B, and 32B scales, together with LLaMA-2-based Emotion-LLaMA models at 7B and 13B scales. For each model, we quantify module-wise parameter changes using a global Frobenius metric, computed between the projection weights of the affectively adapted model and those of its corresponding base model across all layers.

\begin{figure*}[ht]
    \centering

    % ---------------- Row 1 ----------------
    \begin{subfigure}[t]{0.48\linewidth}
        \centering
        \includegraphics[width=\linewidth]{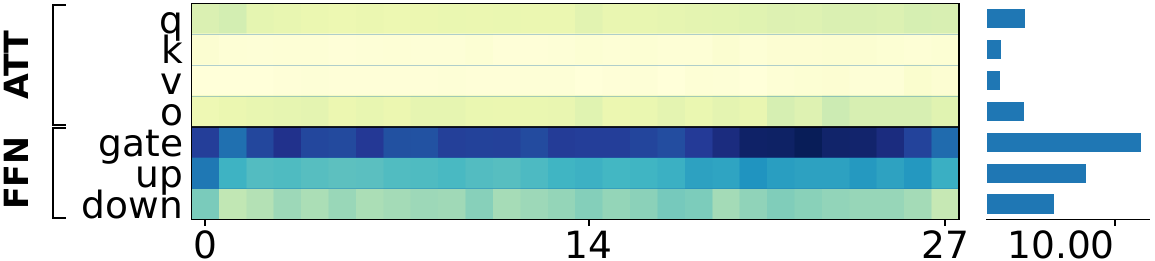}
        \caption{AffectGPT-1.5B}
        \label{fig:affectgpt-1.5b}
    \end{subfigure}
    \hfill
    \begin{subfigure}[t]{0.48\linewidth}
        \centering
        \includegraphics[width=\linewidth]{figures/AffectGPT-7B-with-table_relative_ratio.pdf}
        \caption{AffectGPT-7B}
        \label{fig:affectgpt-7b}
    \end{subfigure}

    \vspace{0.6em}

    % ---------------- Row 2 ----------------
    \begin{subfigure}[t]{0.48\linewidth}
        \centering
        \includegraphics[width=\linewidth]{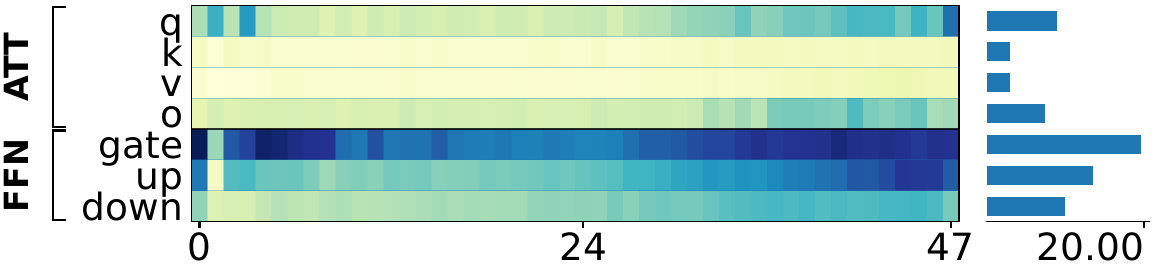}
        \caption{AffectGPT-14B}
        \label{fig:affectgpt-14b}
    \end{subfigure}
    \hfill
    \begin{subfigure}[t]{0.48\linewidth}
        \centering
        \includegraphics[width=\linewidth]{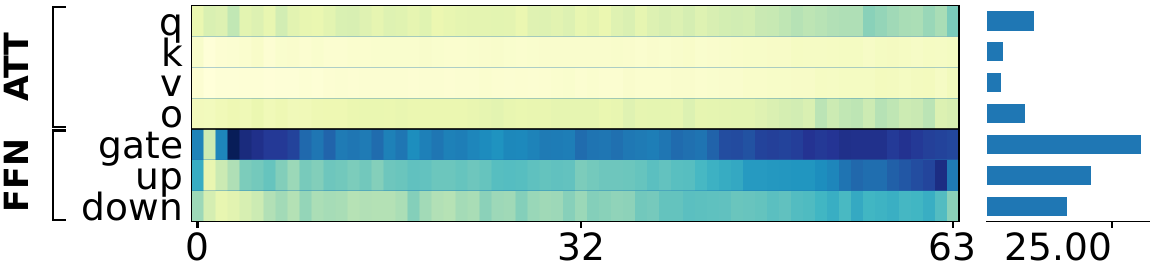}
        \caption{AffectGPT-32B}
        \label{fig:affectgpt-32b}
    \end{subfigure}

    \caption{Layer-wise module weight changes across model families and scales. Across AffectGPT models of different sizes, affective adaptation consistently exhibits an FFN-dominant, non-uniform pattern of parameter changes.}
    \label{fig:cross_scale_weight_changes_affectgpt}
\end{figure*}

\begin{figure*}[ht]
    \centering
    \begin{subfigure}[t]{0.48\linewidth}
        \centering
        \includegraphics[width=\linewidth]{figures/Emotion-LLaMA-7B-with-table_relative_ratio.pdf}
        \caption{Emotion-LLaMA-7B}
        \label{fig:emotion-llama-7b}
    \end{subfigure}
    \hfill
    \begin{subfigure}[t]{0.48\linewidth}
        \centering
        \includegraphics[width=\linewidth]{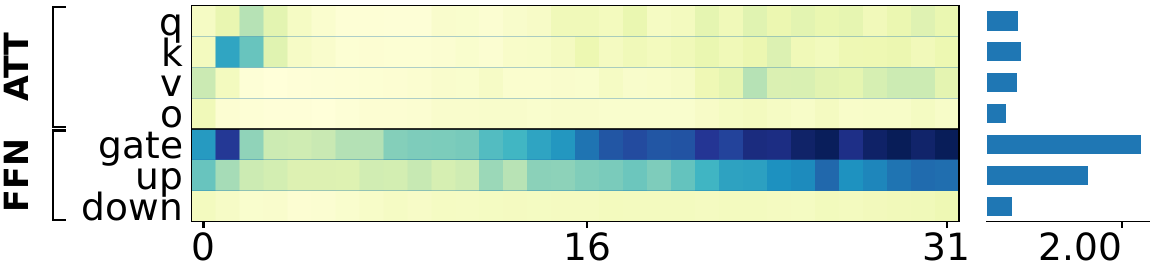}
        \caption{Emotion-LLaMA-13B}
        \label{fig:emotion-llama-13b}
    \end{subfigure}

    \caption{Layer-wise module weight changes across model families and scales. Across Emotion-LLaMA models of different sizes, affective adaptation consistently exhibits an FFN-dominant, non-uniform pattern of parameter changes.}
    \label{fig:cross_scale_weight_changes_emotionllama}
\end{figure*}

As shown in Figure~\ref{fig:cross_scale_weight_changes_affectgpt} and \ref{fig:cross_scale_weight_changes_emotionllama}, a consistent module-level pattern emerges across both model families and scales. Attention projections generally exhibit relatively limited parameter changes, whereas FFN projections undergo substantially more pronounced changes throughout the network. This FFN-dominant pattern is evident from the smallest AffectGPT-1.5B model to the larger 14B and 32B variants, and is similarly observed in both Emotion-LLaMA-7B and Emotion-LLaMA-13B. At the same time, the relative magnitude and layer-wise distribution of changes among individual FFN projections vary across models, indicating that the detailed adaptation pattern is not tied to a particular projection. These results show that the non-uniform organization of affective adaptation is robust across model families and scales. Despite differences in architecture and parameter size, affective adaptation consistently induces stronger changes in the FFN than in Attention, supporting \textbf{the generality of the FFN-dominant module-level adaptation pattern rather than a scale- or architecture-specific effect}.

\subsection{Generality across Affective Tasks and Training Paradigms}
\label{appendix:affective_task_generality}

\begin{figure*}[ht]
    \centering

    % ---------------- Row 1 ----------------
    \begin{subfigure}[t]{0.48\linewidth}
        \centering
        \includegraphics[width=\linewidth]{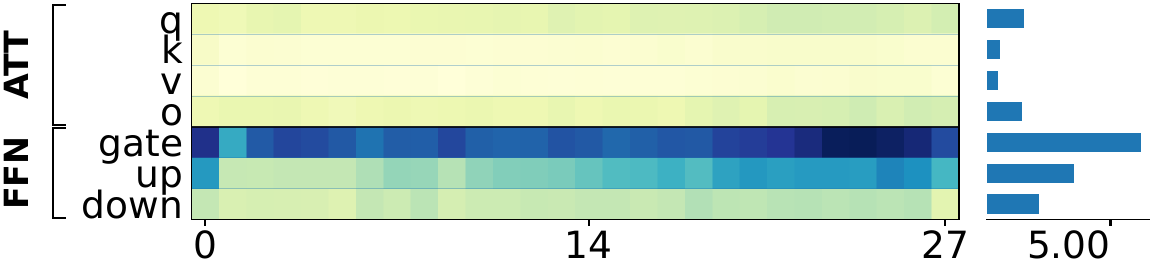}
        \caption{EFT-CoT}
        \label{fig:EFT-COT}
    \end{subfigure}
    \hfill
    \begin{subfigure}[t]{0.48\linewidth}
        \centering
        \includegraphics[width=\linewidth]{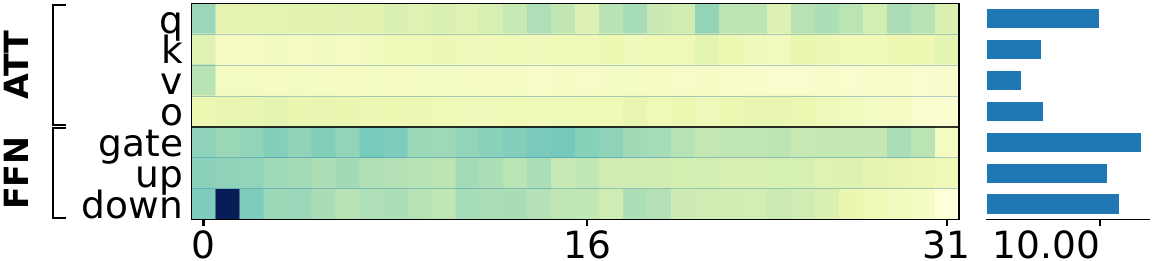}
        \caption{Financial-Sentiment}
        \label{fig:Financial-Sentiment}
    \end{subfigure}

    % ---------------- Row 2 ----------------
    \begin{subfigure}[t]{0.48\linewidth}
        \centering
        \includegraphics[width=\linewidth]{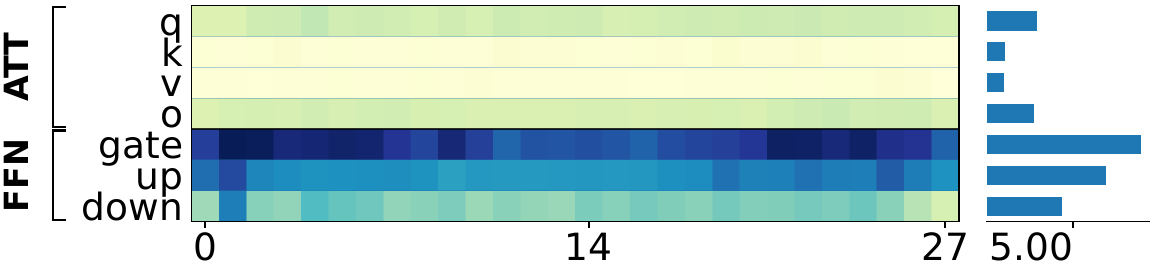}
        \caption{EmotionThinker}
        \label{fig:EmotionThinker}
    \end{subfigure}
    \hfill
    \begin{subfigure}[t]{0.48\linewidth}
        \centering
        \includegraphics[width=\linewidth]{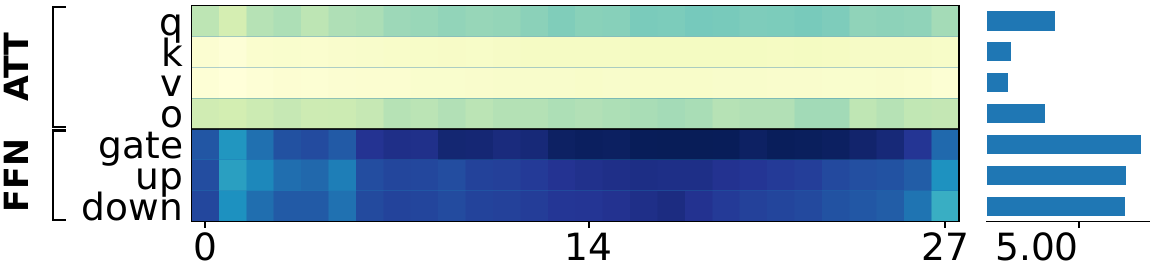}
        \caption{VidEmo}
        \label{fig:videmo}
    \end{subfigure}

    \caption{Visualizations of the affective models not shown in the main text.}

    \label{fig:diff_models}
\end{figure*}

To further examine the generality of the observed module-level adaptation pattern, we extend our analysis to a broader collection of affective models covering diverse tasks and training paradigms. As summarized in Table~\ref{tab:affective_model_coverage}, these models span open-vocabulary emotion recognition, emotion description, emotional support dialogue, multimodal depression recognition, skeleton-based emotion recognition, speech emotion recognition, video emotion analysis, sentiment analysis, empathetic response generation, and therapy-oriented affective reasoning. Their adaptation strategies also vary substantially, including LoRA-based fine-tuning, reinforcement learning, GRPO-based optimization, and full-parameter supervised fine-tuning. We further provide visualizations of the additional representative affective models not shown in the main text, as shown in Figure~\ref{fig:diff_models}, providing additional evidence that the observed module-level adaptation pattern is consistent across diverse model architectures and training paradigms.

\begin{table*}[t]
\centering
\caption{Affective models included in the extended generality analysis. The models cover diverse affective tasks and adaptation paradigms.}
\label{tab:affective_model_coverage}
\small
\renewcommand{\arraystretch}{1.15}
\setlength{\tabcolsep}{5pt}

\begin{tabularx}{\textwidth}{
    >{\raggedright\arraybackslash}p{0.25\textwidth}
    >{\raggedright\arraybackslash}X
    >{\raggedright\arraybackslash}p{0.23\textwidth}
}
\toprule
\textbf{Model} & \textbf{Affective Task} & \textbf{Training Strategy} \\
\midrule

\rowcolor{gray!15}
\multicolumn{3}{l}{\hspace{2pt}\textit{LoRA-based fine-tuning}} \\

AffectGPT
& \makecell[l]{Open-vocabulary emotion recognition \\ 
and emotion description}
& LoRA \\

Emotion-LLaMA
& Emotion recognition
& LoRA \\

SCD-MLLM
& Multimodal depression recognition
& LoRA \\

EAI-LLM
& Skeleton-based emotion recognition
& LoRA \\

EFT-CoT
& \makecell[l]{Emotion-focused therapy and\\
therapeutic-process reasoning}
& LoRA \\

llama-3.1-8b-financial-sentiment
& Three-way financial sentiment classification
& QLoRA \\

\rowcolor{gray!15}
\multicolumn{3}{l}{\hspace{2pt}\textit{Other adaptation paradigms}} \\

ESCRAG-R1
& Emotional support dialogue
& SFT + GRPO \\

EmotionThinker
& Speech emotion recognition
& SFT + LoRA + GRPO \\

VidEmo-T1
& \makecell[l]{Fine-grained video emotion and\\
facial-expression analysis}
& SFT + GRPO \\

\bottomrule
\end{tabularx}
\end{table*}

For each model, we compare the adapted model with its corresponding base model and analyze module-wise parameter changes across Attention and FFN projections following the same protocol used in our main analysis. Since these models differ substantially in architecture, parameter scale, task formulation, and training procedure, we focus on the relative distribution of parameter changes within each model rather than directly comparing their absolute magnitudes.

Across this diverse set of affective models, we consistently observe non-uniform module-level adaptation patterns. Despite substantial differences in affective task formulation and optimization strategy, FFN projections generally exhibit more pronounced changes than Attention projections. This trend is observed not only in conventional emotion recognition and sentiment analysis, but also in affective generation, multimodal depression recognition, speech- and video-based emotion understanding, and affective reasoning tasks.

These results provide additional evidence that the FFN-dominant organization observed in our main experiments is not confined to a particular affective task, modality, or training paradigm. \textbf{Across the examined affective settings, structured and non-uniform parameter changes emerge as a recurring pattern}.

\subsection{Robustness Analysis with Alternative Weight-Change Metrics}
\label{appendix:B}

To further verify that the observed parameter changes are not specific to a particular distance metric, we evaluate the magnitude of weight updates using several complementary statistics. For each module, we define the parameter update as $\Delta W = W_{\mathrm{aff}} - W_{\mathrm{base}}$, where $W_{\mathrm{aff}}$ and $W_{\mathrm{base}}$ denote the weights of the affectively adapted and base models, respectively. Specifically, we consider the mean absolute update,

$$
\mathrm{MAE}(\Delta W)=\frac{1}{N}\sum_{i=1}^{N}|\Delta w_i|,
$$

the median absolute update,

$$
\mathrm{Median}(|\Delta W|),
$$

and the $90$th, and $99$th percentiles of the absolute update distribution,

$$
P_{90}(|\Delta W|),\quad P_{99}(|\Delta W|).
$$

These metrics capture complementary aspects of the parameter-update distribution. MAE measures the average magnitude of parameter changes, while the median reflects the typical update magnitude and is less sensitive to extreme values. The upper percentiles characterize relatively large updates and provide a more robust view of the upper tail of the distribution without relying on a small number of extreme values.

\begin{figure*}[ht]
    \centering

    % ---------------- Row 1 ----------------
    \begin{subfigure}[t]{0.48\linewidth}
        \centering
        \includegraphics[width=\linewidth]{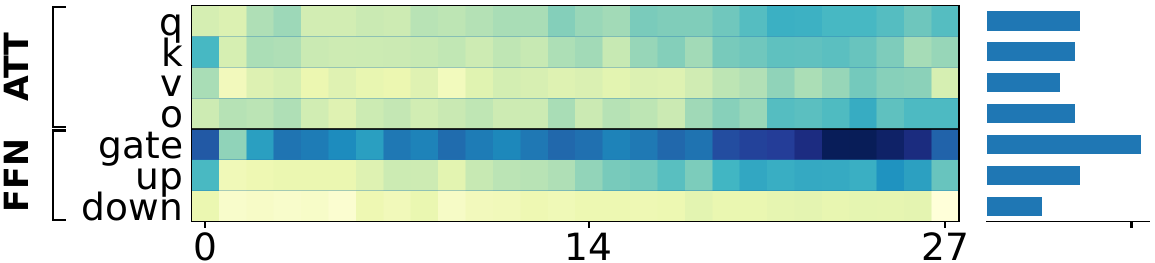}
        \caption{MAE Metrics}
        \label{fig:affectgpt-1.5b}
    \end{subfigure}
    \hfill
    \begin{subfigure}[t]{0.48\linewidth}
        \centering
        \includegraphics[width=\linewidth]{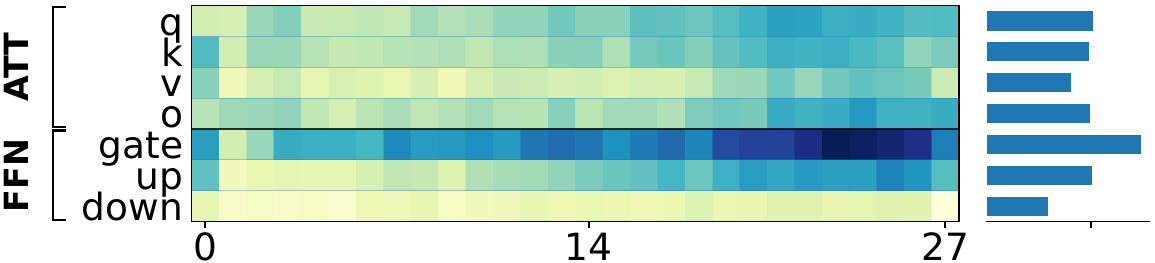}
        \caption{Median Metrics}
        \label{fig:affectgpt-7b}
    \end{subfigure}

    \vspace{0.6em}

    % ---------------- Row 2 ----------------
    \begin{subfigure}[t]{0.48\linewidth}
        \centering
        \includegraphics[width=\linewidth]{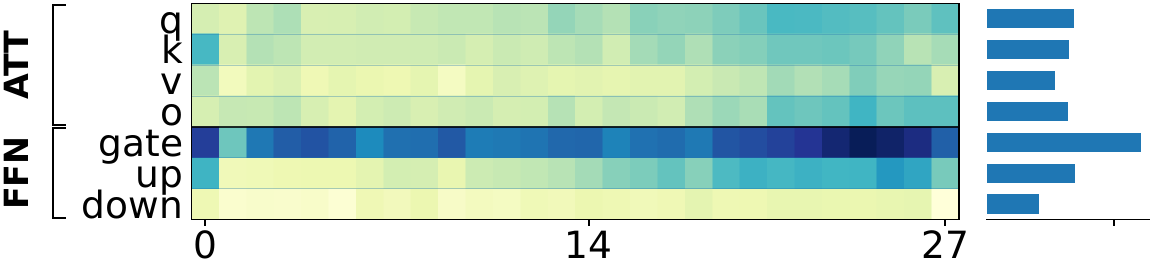}
        \caption{$P_{90}$ Metrics}
        \label{fig:affectgpt-14b}
    \end{subfigure}
    \hfill
    \begin{subfigure}[t]{0.48\linewidth}
        \centering
        \includegraphics[width=\linewidth]{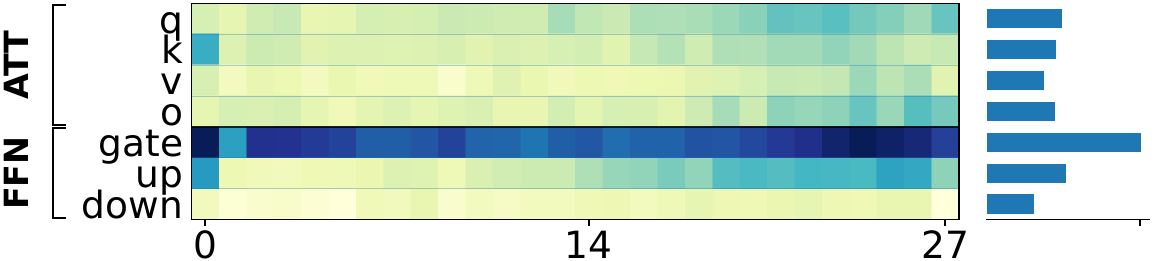}
        \caption{$P_{99}$ Metrics}
        \label{fig:affectgpt-32b}
    \end{subfigure}

    \caption{
Robustness of module-wise parameter changes under alternative weight-change metrics on EFT-CoT.
Panels (a)--(d) report layer-wise parameter changes measured by MAE, median absolute update, $P_{90}$, and $P_{99}$, respectively.
The bar chart in each panel reports the corresponding metric aggregated across layers.
Across these metrics, FFN projections generally exhibit larger parameter changes than Attention projections, indicating that the observed module-wise asymmetry is robust to the choice of weight-change metric.
}
    \label{fig:diff-metric}
\end{figure*}
Across these alternative metrics, a consistent module-wise asymmetry remains evident, as shown in Figure~\ref{fig:diff-metric}. In particular, \texttt{gate\_proj} exhibits the most pronounced parameter changes across MAE, median, $P_{90}$, and $P_{99}$, while \texttt{up\_proj} also shows relatively large updates compared with most Attention projections. Importantly, this pattern appears not only in the upper tail of the update distribution but also in its mean and median. The consistency across multiple distributional statistics suggests that the prominence of \texttt{gate\_proj} is not driven solely by a small number of extreme parameter updates, but instead reflects a broader shift in its parameter distribution during affective adaptation. These results further support the robustness of the non-uniform adaptation pattern observed across Transformer modules.

\begin{table*}[t]

\centering

\caption{Non-affective models included in the extended generality analysis.
The models cover diverse tasks and post-training paradigms.}

\label{tab:non_affective_model_coverage}

\small

\renewcommand{\arraystretch}{1.15}

\setlength{\tabcolsep}{5pt}

\begin{tabularx}{\textwidth}{
    >{\raggedright\arraybackslash}p{0.26\textwidth}
    >{\raggedright\arraybackslash}X
    >{\raggedright\arraybackslash}p{0.22\textwidth}
}

\toprule

\textbf{Model}
& \textbf{Task}
& \textbf{Training Strategy} \\

\midrule

\rowcolor{gray!15}
\multicolumn{3}{l}{
    \hspace{2pt}\textit{Instruction fine-tuning}
} \\

\makecell[l]{deepseek-coder-1.3b\\-instruct}
& \makecell[l]{Code generation, code completion,\\
programming question answering, and\\
instruction following}
& Instruction SFT \\

\makecell[l]{SmolLM-1.7B-Instruct\\(v0.2)}
& \makecell[l]{English general question answering,\\
daily dialogue, creative writing, and\\
basic Python programming}
& SFT \\

\rowcolor{gray!15}
\multicolumn{3}{l}{
    \hspace{2pt}\textit{Reasoning and capability distillation}
} \\

\makecell[l]{DeepSeek-R1-Distill\\-Qwen-1.5B/7B/14B}
& \makecell[l]{Mathematical, coding, and general\\
reasoning with long reasoning chains}
& \makecell[l]{Knowledge distillation\\via SFT} \\

\makecell[l]{Qwen3-0.6B}
& \makecell[l]{Controllable thinking and non-thinking,\\
mathematical, coding, and logical reasoning,\\
instruction following, tool use, and\\
multilingual generation}
& \makecell[l]{Strong-to-weak distillation\\
(off-policy $\rightarrow$ on-policy)} \\

\rowcolor{gray!15}
\multicolumn{3}{l}{
    \hspace{2pt}\textit{Preference optimization and reinforcement learning}
} \\

\makecell[l]{Qwen2.5-Coder-1.5B\\-Instruct}
& \makecell[l]{Code generation, code reasoning,\\
code repair, and programming\\
instruction following}
& SFT $\rightarrow$ DPO \\

\makecell[l]{granite-3.2-2b\\-instruct}
& \makecell[l]{Controllable reasoning, instruction\\
following, question answering, summarization,\\
RAG, coding, and tool use}
& TPO-based RL \\

\makecell[l]{OLMo-2-0425-1B-DPO}
& \makecell[l]{General dialogue, instruction following,\\
response-preference alignment, and\\
mathematical tasks}
& DPO \\

\makecell[l]{OLMo-2-0425-1B\\-RLVR1}
& \makecell[l]{Mathematical reasoning and verifiable\\
instruction following under mixed GSM,\\
MATH, and IF constraints}
& RLVR \\

\bottomrule

\end{tabularx}

\end{table*}

\subsection{Comparison with Non-Affective Adaptation}
\label{appendix:Non-Affective Adaptation}
To provide an additional comparison, we extend the module-level analysis to several non-affective models spanning question answering, code generation, mathematical reasoning, and logical reasoning, as shown in Tab~\ref{tab:non_affective_model_coverage}. These models cover different post-training paradigms, including instruction fine-tuning, knowledge distillation, preference optimization, and reinforcement learning. We apply the same module-wise analysis to examine how parameter changes are distributed across Attention and FFN projections.

% 与情感/非情感模型进行全面对比
\begin{figure*}[ht]
  \centering

  % =========================================================
  % Affective Tasks
  % =========================================================

  \begin{minipage}[b]{0.49\linewidth}
    \centering
    \includegraphics[width=\linewidth]{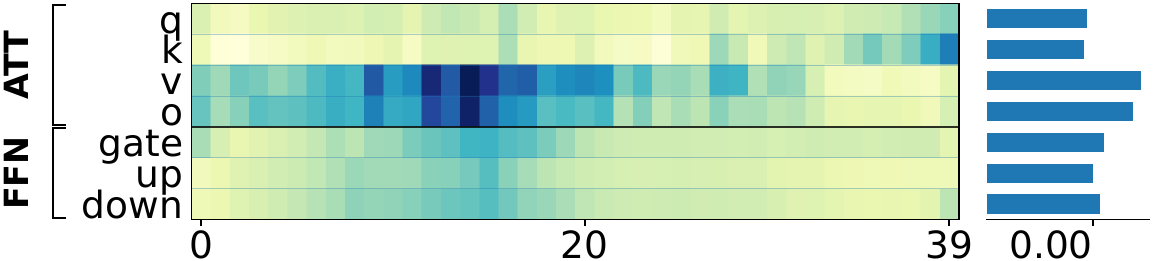}
    {\small (a) Question answering}
  \end{minipage}
  \hfill
  \begin{minipage}[b]{0.49\linewidth}
    \centering
    \includegraphics[width=\linewidth]{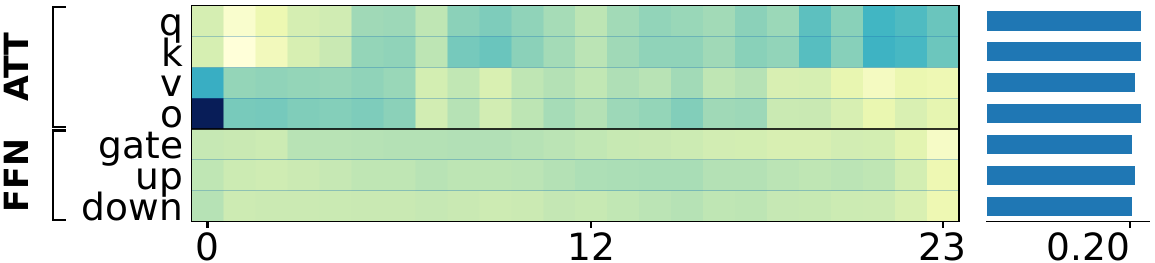}
    {\small (b) Code generation}
  \end{minipage}

  \vspace{0.9em}

  % =========================================================
  % Non-Affective Tasks
  % =========================================================

  \begin{minipage}[b]{0.49\linewidth}
    \centering
    \includegraphics[width=\linewidth]{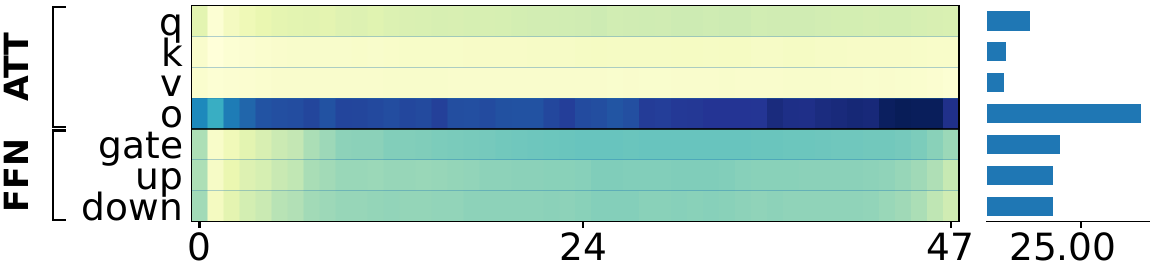}
    {\small (c) Mathematical Reasoning}
  \end{minipage}
  \hfill
  \begin{minipage}[b]{0.49\linewidth}
    \centering
    \includegraphics[width=\linewidth]{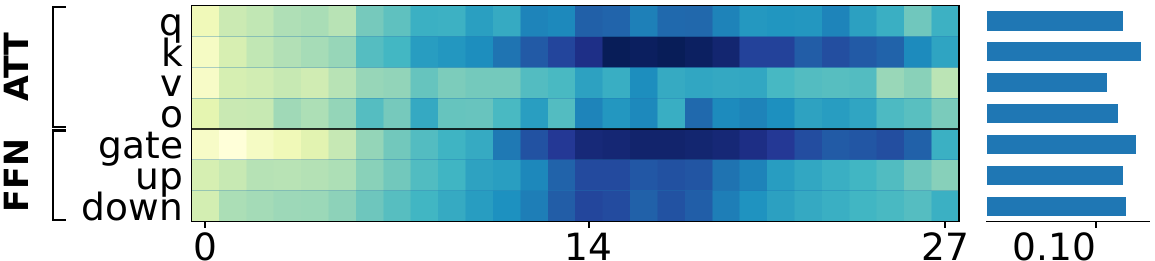}
    {\small (d) Logical Reasoning}
  \end{minipage}
  \caption{
Module-level parameter changes across non-affective tasks, including (a) question answering (granite-3.2-2b-instruct), (b) code generation (deepseek-coder-1.3b-instruct), (c) mathematical reasoning (DeepSeek-R1-Distill-Qwen-14B), and (d) logical reasoning (Qwen3-06B).
The heatmaps show layer-wise parameter changes across Transformer projections, with the horizontal axis indicating layer indices; the bars on the right report the mean L2 difference of each projection across layers.
The examined tasks exhibit diverse module-level patterns: question answering and mathematical reasoning show relatively pronounced changes in \texttt{o\_proj}, whereas code generation and logical reasoning display more distributed changes across projections and layers.
}
  \label{fig:3.3}
\end{figure*}

As shown in Figure~\ref{fig:3.3}, the examined non-affective settings exhibit diverse module-level patterns. For example, question answering and mathematical reasoning show relatively pronounced changes in \texttt{o\_proj}, whereas code generation and logical reasoning display more distributed changes across multiple projections and layers. Overall, both the dominant projections and their layer-wise distributions vary across the examined settings.
These results indicate that the module-level patterns observed in the non-affective models differ from those identified in the affective models. In particular, the consistent FFN-dominant tendency observed across the examined affective settings is not reproduced uniformly in these non-affective cases. Although the current comparisons are insufficient to establish task-specific functional organization, they provide additional evidence that the observed FFN-dominant pattern is not simply shared across all forms of post-training.

\subsection{Training Dynamics across Additional Models}
\label{appendix:training_dynamics}

\begin{figure*}[ht]
  \centering
  \includegraphics[width=0.5\linewidth]{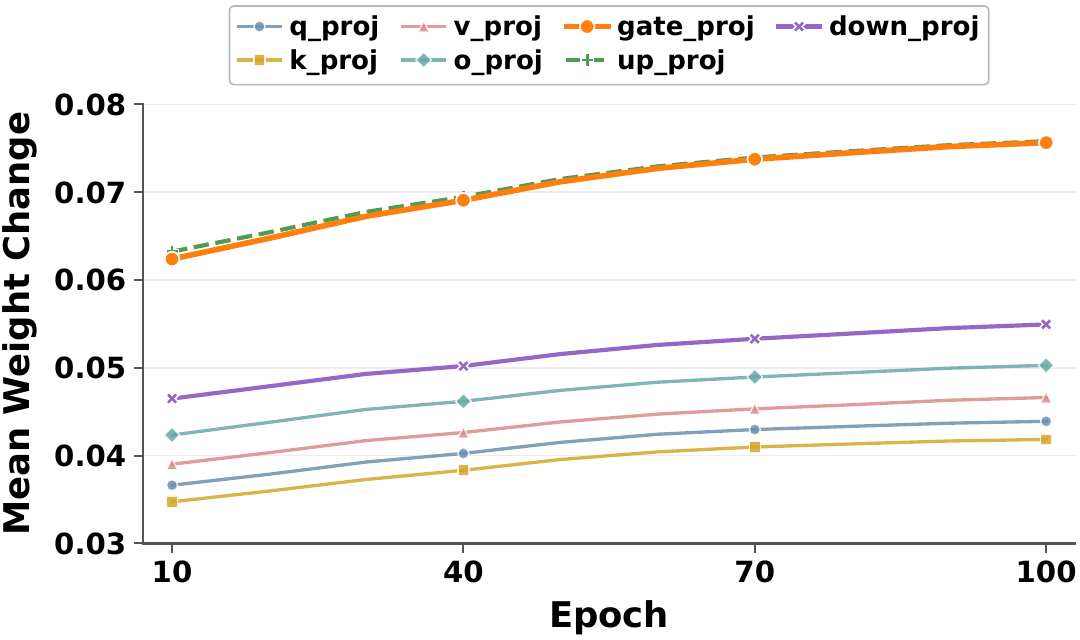}
  \caption{
Training dynamics of projection-wise LoRA update magnitudes in EAI-LLM.
Different projection pathways exhibit non-uniform optimization trajectories, with \texttt{gate\_proj} and \texttt{up\_proj} showing similarly large updates, while \texttt{down\_proj} and the Attention projections follow distinct trajectories.
}
  \label{fig:EAI-LLM_trained}
\end{figure*}

To examine whether the parameter-level differentiation observed during training is specific to AffectGPT, we additionally track projection-wise LoRA update magnitudes in EAI-LLM~\citep{lu2025understanding}. As shown in Figure~\ref{fig:EAI-LLM_trained}, the pathways again exhibit clearly non-uniform optimization trajectories. In this model, \texttt{gate\_proj} and \texttt{up\_proj} show similarly large updates, while \texttt{down\_proj} and the Attention projections follow distinct trajectories. These results suggest that pathway-level differentiation during adaptation is not limited to a single model, although the relative ordering among FFN pathways may vary across architectures.

\subsection{Additional Analysis of Module-wise Functional Contributions}
\label{appendix:E}

This section provides additional experiments to further characterize the
functional contributions of different Transformer pathways during affective
adaptation. We consider two complementary settings. First, we examine how
LoRA updates learned during full affective adaptation interact when they are
selectively recovered and combined in the base model. Second, we retrain
individual modules or module combinations under controlled settings to
evaluate their independent and collaborative adaptation capacity. Together,
these experiments distinguish the functional specialization observed in the
final adapted model from the intrinsic adaptation capacity of individual
modules.

\subsubsection{Compositional Analysis of Recovered LoRA Updates}
\textbf{Experiment 1: Comparison among FFN pathways}

We first examine whether the stronger contribution of the gating pathway
observed after full affective adaptation is shared by other structurally
similar FFN pathways. Specifically, we combine the recovered LoRA weights of
each Attention projection with those of \texttt{gate\_proj},
\texttt{up\_proj}, or \texttt{down\_proj}, and load each combination into the
same base model.

\begin{table}[ht]
    \centering
    \caption{Compositional contributions of recovered FFN LoRA updates.
    Each cell reports the resulting mean performance, with the absolute change
    relative to the corresponding Attention-only baseline shown in parentheses.}
    \label{tab:appendix_ffn_recovery}
    \small
    \setlength{\tabcolsep}{7pt}
    \renewcommand{\arraystretch}{1.15}

    \begin{tabular}{lcccc}
        \toprule
        Base Module
        & Base
        & $+\texttt{gate\_proj}$
        & $+\texttt{up\_proj}$
        & $+\texttt{down\_proj}$ \\
        \midrule

        \texttt{q\_proj}
        & 58.72
        & \cellcolor{gray!10}\textbf{65.55 {\scriptsize($+6.83$)}}
        & 61.27 {\scriptsize($+2.55$)}
        & 61.60 {\scriptsize($+2.88$)} \\

        \texttt{k\_proj}
        & 60.57
        & \cellcolor{gray!10}\textbf{63.98 {\scriptsize($+3.41$)}}
        & 61.40 {\scriptsize($+0.83$)}
        & 60.80 {\scriptsize($+0.23$)} \\

        \texttt{v\_proj}
        & 57.76
        & \cellcolor{gray!10}\textbf{64.12 {\scriptsize($+6.36$)}}
        & 58.92 {\scriptsize($+1.16$)}
        & 58.76 {\scriptsize($+1.00$)} \\

        \texttt{o\_proj}
        & 59.93
        & \cellcolor{gray!10}\textbf{62.62 {\scriptsize($+2.69$)}}
        & 60.12 {\scriptsize($+0.19$)}
        & 59.31 {\scriptsize($-0.62$)} \\

        \bottomrule
    \end{tabular}
\end{table}

As shown in Table~\ref{tab:appendix_ffn_recovery}, adding the recovered
\texttt{gate\_proj} updates consistently produces larger performance gains
across different Attention projections than adding \texttt{up\_proj} or
\texttt{down\_proj}. This suggests that, in the fully adapted model, the
updates learned by the gating pathway make a stronger compositional
contribution to affective performance.

\textbf{Experiment 2: Cross-structure composition with Attention pathways}

We next compare combinations formed by two Attention pathways with those
formed by an Attention pathway and \texttt{gate\_proj}. Starting from the
performance obtained by recovering a single Attention projection, we
separately add another Attention projection or the recovered
\texttt{gate\_proj} updates.

\begin{table*}[ht]
    \centering
    \caption{Comparison of recovered LoRA update compositions across Attention
    and gating pathways. Each cell reports the resulting mean performance, with
    the absolute change relative to the corresponding Attention-only baseline
    shown in parentheses.}
    \label{tab:appendix_cross_structure_recovery}
    \small
    \setlength{\tabcolsep}{7pt}
    \renewcommand{\arraystretch}{1.15}

    \begin{tabular}{lccccc}
        \toprule
        Base Module
        & Base
        & $+\texttt{q\_proj}$
        & $+\texttt{k\_proj}$
        & $+\texttt{v\_proj}$
        & $+\texttt{gate\_proj}$ \\
        \midrule

        \texttt{q\_proj}
        & 58.72
        & --
        & 60.09 {\scriptsize($+1.37$)}
        & 58.68 {\scriptsize($-0.04$)}
        & \cellcolor{gray!10}\textbf{65.55 {\scriptsize($+6.83$)}} \\

        \texttt{k\_proj}
        & 60.57
        & 60.09 {\scriptsize($-0.48$)}
        & --
        & 58.96 {\scriptsize($-1.61$)}
        & \cellcolor{gray!10}\textbf{63.98 {\scriptsize($+3.41$)}} \\

        \texttt{v\_proj}
        & 57.76
        & 58.68 {\scriptsize($+0.92$)}
        & 58.96 {\scriptsize($+1.20$)}
        & --
        & \cellcolor{gray!10}\textbf{64.12 {\scriptsize($+6.36$)}} \\

        \texttt{o\_proj}
        & 59.93
        & \textbf{62.63 {\scriptsize($+2.70$)}}
        & 61.36 {\scriptsize($+1.43$)}
        & 60.40 {\scriptsize($+0.47$)}
        & \cellcolor{gray!10}62.62 {\scriptsize($+2.69$)} \\

        \bottomrule
    \end{tabular}
\end{table*}

Table~\ref{tab:appendix_cross_structure_recovery} shows that combinations
between Attention projections generally provide smaller and less consistent
gains, whereas adding \texttt{gate\_proj} produces more stable improvements.
These results indicate that the contribution of the recovered gating updates
cannot be explained simply by combining additional adapted modules.

\textbf{Experiment 3: Marginal contribution under stronger Attention recovery}

We further test whether this contribution persists when a larger fraction of
the learned Attention updates is already recovered. We first load pairs of
Attention LoRA updates into the base model and then additionally recover the
LoRA weights of \texttt{gate\_proj}, \texttt{up\_proj}, or
\texttt{down\_proj}.

\begin{table}[ht]
    \centering
    \caption{Marginal contributions of recovered FFN LoRA updates under stronger
    Attention recovery. Each cell reports the resulting mean performance, with
    the absolute change relative to the corresponding Attention-pair baseline
    shown in parentheses.}
    \label{tab:appendix_attention_pair_recovery}
    \small
    \setlength{\tabcolsep}{6pt}
    \renewcommand{\arraystretch}{1.15}

    \begin{tabular}{lcccc}
        \toprule
        Attention Pair
        & Base
        & $+\texttt{gate\_proj}$
        & $+\texttt{up\_proj}$
        & $+\texttt{down\_proj}$ \\
        \midrule

        \texttt{q+k}
        & 60.09
        & \cellcolor{gray!10}\textbf{66.47 {\scriptsize($+6.38$)}}
        & 62.39 {\scriptsize($+2.30$)}
        & 63.04 {\scriptsize($+2.95$)} \\

        \texttt{q+v}
        & 58.68
        & \cellcolor{gray!10}\textbf{65.53 {\scriptsize($+6.85$)}}
        & 62.25 {\scriptsize($+3.57$)}
        & 61.30 {\scriptsize($+2.62$)} \\

        \texttt{q+o}
        & 62.63
        & \cellcolor{gray!10}\textbf{64.21 {\scriptsize($+1.58$)}}
        & 61.86 {\scriptsize($-0.77$)}
        & 61.41 {\scriptsize($-1.22$)} \\

        \texttt{k+v}
        & 58.96
        & \cellcolor{gray!10}\textbf{65.85 {\scriptsize($+6.89$)}}
        & 62.01 {\scriptsize($+3.05$)}
        & 61.91 {\scriptsize($+2.95$)} \\

        \texttt{k+o}
        & 61.36
        & \cellcolor{gray!10}\textbf{62.32 {\scriptsize($+0.96$)}}
        & 60.77 {\scriptsize($-0.59$)}
        & 61.02 {\scriptsize($-0.34$)} \\

        \texttt{v+o}
        & 60.40
        & \cellcolor{gray!10}\textbf{62.42 {\scriptsize($+2.02$)}}
        & 60.38 {\scriptsize($-0.02$)}
        & 58.11 {\scriptsize($-2.29$)} \\

        \bottomrule
    \end{tabular}
\end{table}

As shown in Table~\ref{tab:appendix_attention_pair_recovery},
\texttt{gate\_proj} continues to provide a relatively stable positive gain
across different Attention combinations, whereas the gains from
\texttt{up\_proj} and \texttt{down\_proj} are less consistent. This result
further supports the pronounced functional contribution of the gating
pathway in the final affectively adapted model.

\subsubsection{Controlled Analysis of Module Adaptation Capacity}
The recovery experiments above characterize the functional organization of
updates learned during full affective adaptation. However, they do not
establish whether the observed differences arise from intrinsic properties of
the modules or from the optimization process. We therefore conduct additional
controlled training experiments.

\textbf{Experiment 1: Parameter-matched single-module adaptation}

We first control for differences in trainable parameter count and independently
adapt each projection while keeping the remaining model parameters frozen.
The LoRA configurations are adjusted such that the numbers of trainable
parameters are approximately matched across modules.

\begin{table*}[t]
    \centering
    \caption{
Performance comparison of single-module adaptation under approximately matched trainable-parameter budgets across all seven projection modules.
\emph{Drop} denotes the absolute decrease in mean performance relative to AffectGPT.
Higher Mean and lower Drop indicate better performance preservation.
}
    \label{tab:appendix_parameter_matched}

    \small
    \setlength{\tabcolsep}{3.2pt}
    \renewcommand{\arraystretch}{1.10}

    \begin{adjustbox}{max width=\textwidth}
    \begin{tabular}{llcccccccccc}
        \toprule

        \multirow{2}{*}{Group}
        & \multirow{2}{*}{Module}
        & \multicolumn{4}{c}{Emotion Recognition}
        & \multicolumn{4}{c}{Sentiment Analysis}
        & \multicolumn{2}{c}{Overall} \\

        \cmidrule(lr){3-6}
        \cmidrule(lr){7-10}
        \cmidrule(lr){11-12}

        &
        & MER2023
        & MER2024
        & MELD
        & IEMOCAP
        & MOSI
        & MOSEI
        & SIMS
        & SIMS v2
        & Mean $\uparrow$
        & Drop $\downarrow$ \\

        \midrule

        Reference
        & AffectGPT
        & 78.54
        & 78.80
        & 55.65
        & 60.54
        & 81.30
        & 80.90
        & 88.49
        & 86.18
        & 76.30
        & -- \\

        \midrule

        \multirow{4}{*}{Attention}
        & \texttt{q\_proj}
        & 55.33 & 53.99 & 49.94 & 42.87
        & 76.40 & 63.36 & 81.88 & 81.25
        & 63.13 & 13.17 \\

        & \texttt{k\_proj}
        & 60.35 & 61.39 & 55.87 & 49.01
        & 79.55 & 70.69 & 77.35 & 79.27
        & 66.68 & 9.62 \\

        & \texttt{v\_proj}
        & 52.34 & 56.50 & 45.94 & 46.32
        & 81.25 & 71.93 & 78.43 & 76.07
        & 63.60 & 12.70 \\

        & \texttt{o\_proj}
        & 59.23 & 60.84 & 49.65 & 49.80
        & 81.85 & 64.66 & 81.43 & 80.49
        & 65.99 & 10.31 \\

        \midrule

        \multirow{3}{*}{FFN}
        & \texttt{gate\_proj}
        & 71.63 & 76.11 & 56.48 & 58.69
        & 80.88 & 78.05 & 84.42 & 83.22
        & \cellcolor{gray!15}\textbf{73.68}
        & \cellcolor{gray!15}\textbf{2.62} \\

        & \texttt{up\_proj}
        & 70.78 & 75.30 & 52.85 & 57.55
        & 75.51 & 79.09 & 85.76 & 83.44
        & 72.53 & 3.77 \\

        & \texttt{down\_proj}
        & 75.42 & 74.91 & 54.58 & 57.41
        & 75.73 & 78.25 & 87.77 & 83.34
        & 73.43 & 2.87 \\

        \bottomrule
    \end{tabular}
    \end{adjustbox}
\end{table*}

As shown in Table~\ref{tab:appendix_parameter_matched}, FFN projections
consistently outperform Attention projections under comparable parameter
budgets. Although \texttt{gate\_proj} achieves the highest mean performance,
\texttt{up\_proj} and \texttt{down\_proj} also exhibit strong independent
adaptation capability. These results suggest that the advantage of the FFN is
not simply attributable to differences in trainable parameter count.

\textbf{Experiment 2: Collaborative adaptation among FFN pathways}

We next examine whether \texttt{gate\_proj} retains a unique advantage when
different FFN pathways are jointly trained with the same Attention projection.
For each Attention projection, we separately pair it with
\texttt{gate\_proj}, \texttt{up\_proj}, or \texttt{down\_proj} and train the
corresponding module pair under the same setting.

\begin{table*}[ht]
    \centering
    \caption{Collaborative adaptation performance with different FFN pathways.
    Each cell reports the resulting mean performance, with the absolute gain
    over the corresponding single-module baseline shown in parentheses.}
    \label{tab:appendix_ffn_collaboration}
    \small
    \setlength{\tabcolsep}{10pt}
    \renewcommand{\arraystretch}{1.15}

    \begin{tabular}{lcccc}
        \toprule
        Base Module
        & Base
        & $+\texttt{gate\_proj}$
        & $+\texttt{up\_proj}$
        & $+\texttt{down\_proj}$ \\
        \midrule

        \texttt{q\_proj}
        & 68.98
        & 72.74 {\scriptsize($+3.76$)}
        & \textbf{74.44 {\scriptsize($+5.46$)}}
        & 71.88 {\scriptsize($+2.90$)} \\

        \texttt{k\_proj}
        & 71.89
        & 73.55 {\scriptsize($+1.66$)}
        & 73.74 {\scriptsize($+1.85$)}
        & \textbf{73.77 {\scriptsize($+1.88$)}} \\

        \texttt{v\_proj}
        & 61.98
        & \textbf{74.07 {\scriptsize($+12.09$)}}
        & 73.78 {\scriptsize($+11.80$)}
        & 73.77 {\scriptsize($+11.79$)} \\

        \texttt{o\_proj}
        & 65.66
        & 73.67 {\scriptsize($+8.01$)}
        & \textbf{74.31 {\scriptsize($+8.65$)}}
        & 71.85 {\scriptsize($+6.19$)} \\
        \bottomrule
    \end{tabular}
\end{table*}

As shown in Table~\ref{tab:appendix_ffn_collaboration}, all three FFN
projections consistently improve affective performance when jointly trained
with Attention, and their overall gains are comparable. In contrast to the
recovery results, \texttt{gate\_proj} does not exhibit a stable exclusive
advantage in this controlled training setting. This result suggests that the
stronger gate preference observed after full adaptation is unlikely to arise
solely from an intrinsic advantage of the gating pathway.

\textbf{Experiment 3: Cross-structure collaborative adaptation}

We further compare collaboration between two Attention projections with
collaboration between an Attention projection and \texttt{gate\_proj}. Each
module pair is jointly trained from the same base model under identical
training conditions.

\begin{table*}[t]
    \centering
    \caption{Cross-structure collaborative adaptation between Attention and gating pathways. Each cell reports the resulting mean performance, with the absolute gain over the corresponding single-module baseline shown in parentheses.}
    \label{tab:appendix_cross_structure_training}

    \small
    \setlength{\tabcolsep}{7pt}
    \renewcommand{\arraystretch}{1.15}

    \begin{adjustbox}{max width=\textwidth}
    \begin{tabular}{lcccccc}
        \toprule
        Base Module
        & Base
        & $+\texttt{q\_proj}$
        & $+\texttt{k\_proj}$
        & $+\texttt{v\_proj}$
        & $+\texttt{o\_proj}$
        & $+\texttt{gate\_proj}$ \\
        \midrule

        \texttt{q\_proj}
        & 68.98
        & --
        & 72.49 {\scriptsize($+3.51$)}
        & \textbf{72.90 {\scriptsize($+3.92$)}}
        & 72.82 {\scriptsize($+3.84$)}
        & 72.74 {\scriptsize($+3.76$)} \\

        \texttt{k\_proj}
        & 71.89
        & 72.49 {\scriptsize($+0.60$)}
        & --
        & \textbf{73.89 {\scriptsize($+2.00$)}}
        & 72.69 {\scriptsize($+0.80$)}
        & 73.55 {\scriptsize($+1.66$)} \\

        \texttt{v\_proj}
        & 61.98
        & 72.90 {\scriptsize($+10.92$)}
        & 73.89 {\scriptsize($+11.91$)}
        & --
        & 73.65 {\scriptsize($+11.67$)}
        & \textbf{74.07 {\scriptsize($+12.09$)}} \\

        \texttt{o\_proj}
        & 65.66
        & 72.82 {\scriptsize($+7.16$)}
        & 72.69 {\scriptsize($+7.03$)}
        & 73.65 {\scriptsize($+7.99$)}
        & --
        & \textbf{73.67 {\scriptsize($+8.01$)}} \\

        \bottomrule
    \end{tabular}
    \end{adjustbox}
\end{table*}

Table~\ref{tab:appendix_cross_structure_training} shows that both
Attention--Attention and Attention--gate combinations substantially improve
affective performance. Although combinations involving \texttt{gate\_proj}
achieve a slightly higher average gain, this advantage is not consistently
observed across all module pairs. These results suggest that both
within-structure and cross-structure module combinations can effectively
support affective adaptation.

\textbf{Experiment 4: Marginal FFN contribution under enhanced Attention adaptation}

Finally, we examine whether FFN pathways continue to provide additional
affective adaptation capacity when the Attention side has already been
strengthened. We first jointly train pairs of Attention projections and then
add \texttt{gate\_proj}, \texttt{up\_proj}, or \texttt{down\_proj} to the
training configuration.

\begin{table*}[ht]
    \centering
    \caption{Marginal contributions of FFN pathways under stronger Attention
    adaptation. Each cell reports the resulting mean performance, with the
    absolute gain over the corresponding Attention-pair baseline shown in
    parentheses.}
    \label{tab:appendix_enhanced_attention}
    \small
    \setlength{\tabcolsep}{10pt}
    \renewcommand{\arraystretch}{1.15}

    \begin{tabular}{lcccc}
        \toprule
        Attention Pair
        & Base
        & $+\texttt{gate\_proj}$
        & $+\texttt{up\_proj}$
        & $+\texttt{down\_proj}$ \\
        \midrule

        \texttt{q+k}
        & 72.49
        & \textbf{74.69 {\scriptsize($+2.20$)}}
        & 73.30 {\scriptsize($+0.81$)}
        & 71.08 {\scriptsize($-1.41$)} \\

        \texttt{q+v}
        & 72.90
        & \textbf{74.32 {\scriptsize($+1.42$)}}
        & 73.05 {\scriptsize($+0.15$)}
        & 71.55 {\scriptsize($-1.35$)} \\

        \texttt{q+o}
        & 72.82
        & \textbf{74.41 {\scriptsize($+1.59$)}}
        & 73.52 {\scriptsize($+0.70$)}
        & 73.39 {\scriptsize($+0.57$)} \\

        \texttt{k+v}
        & 73.89
        & 73.30 {\scriptsize($-0.59$)}
        & 72.42 {\scriptsize($-1.47$)}
        & \textbf{73.96 {\scriptsize($+0.07$)}} \\

        \texttt{k+o}
        & 72.69
        & 73.59 {\scriptsize($+0.90$)}
        & \textbf{74.06 {\scriptsize($+1.37$)}}
        & 73.57 {\scriptsize($+0.88$)} \\

        \texttt{v+o}
        & 73.65
        & 73.81 {\scriptsize($+0.16$)}
        & \textbf{74.54 {\scriptsize($+0.89$)}}
        & 73.89 {\scriptsize($+0.24$)} \\

        \bottomrule
    \end{tabular}
\end{table*}

As shown in Table~\ref{tab:appendix_enhanced_attention}, the marginal gains
from adding FFN pathways generally decrease as the Attention configuration
becomes stronger. Nevertheless, \texttt{gate\_proj} achieves the highest
average gain and remains positive across most combinations. This result
suggests that the gating pathway retains relatively stable complementary
capacity, although its advantage becomes less pronounced as more pathways
participate in joint adaptation.

\subsection{Downstream Dataset Overview}
\label{appendix:C}
Across our experiments, we use 13 datasets spanning affective and non-affective tasks. The affective datasets include MER2023, MER2024, IEMOCAP, MELD, CMU-MOSI, CMU-MOSEI, CH-SIMS, CH-SIMS v2, Emilya, KDAE, and EGBM, covering multimodal emotion recognition, sentiment analysis, and body-motion-based emotion recognition. We additionally use GSM8K and MATH for mathematical reasoning. Brief descriptions of these datasets are provided below.

\textbf{MER2023} \citep{10.1145/3581783.3612836} and \textbf{MER2024} \citep{10.1145/3689092.3689959} are widely used benchmarks for Chinese multimodal emotion recognition, where MER2024 extends MER2023. Both datasets are derived from movies and TV series and are segmented using multiple techniques to ensure that each clip contains a single speaker with relatively complete speech content. They annotate six emotion categories: worry, happiness, neutral, anger, surprise, and sadness.

\textbf{IEMOCAP} \citep{2008IEMOCAP} is one of the most widely used datasets for multimodal emotion recognition. It comprises five sessions, each featuring a male and a female actor performing in a laboratory setting. The dataset includes the following emotion labels: anger, happiness, sadness, neutral, excitement, frustration, fear, surprise, and other. Following previous research~\citep{poria2017context}, in our experiments, we merge surprise into the happiness category.

\textbf{MELD} \citep{poria-etal-2019-meld} extends the text-centric EmotionLines dataset by incorporating audio and visual modalities. The data are sourced from the TV series Friends and annotated with seven emotion categories: anger, joy, sadness, neutral, disgust, fear, and surprise.

\textbf{CMU-MOSI} \citep{zadeh-etal-2017-tensor} and \textbf{CMU-MOSEI} \citep{bagher-zadeh-etal-2018-multimodal} consist of opinion videos collected from online platforms, with CMU-MOSEI serving as an expanded version of CMU-MOSI in both scale and topical diversity. Each sample is annotated with a sentiment intensity score in $[-3, 3]$, where $-3$ denotes extremely negative sentiment and $+3$ denotes extremely positive sentiment.

\textbf{CH-SIMS} \citep{yu-etal-2020-ch} and \textbf{CH-SIMS v2} \citep{liu2022make} focus on affective expressions grounded in Chinese cultural contexts. The data are collected from movies, TV series, and variety shows. Both datasets provide sentiment intensity annotations in $[-1, 1]$, where $-1$ indicates extremely negative affect and $+1$ indicates extremely positive affect.

\textbf{Emilya}~\citep{Fourati_Pelachaud_2018} comprises 8,206 samples. Eleven actors performed eight emotions: Neutral, Joy, Anger, Panic, Fear, Anxiety, Sadness, and Shame, across seven daily actions, including sitting, walking, and lifting. Data were recorded using the Xsens MVN system, capturing 28 3D joints at a frequency of 120 Hz.

\textbf{KDAE}~\citep{sapinski2019multimodal} was collected using a portable motion capture system that tracked 72 body markers at a frequency of 125 Hz. This dataset includes 1,402 recordings of seven emotions: Happiness, Sadness, Neutral, Anger, Disgust, Fear, and Surprise, performed by 22 semi-professional actors. For our analysis, we excluded hand joints and focused on 24 full-body joints.

\textbf{EGBM}~\citep{Zhang_Yu_Zhang_Du_Zhan_Chen_Jiang_Guo_Zhao_Wang_etal._2020} contains 560 samples captured by the Kinect V2 camera, each with a frame rate of 30 Hz. The data included 16 professional actors performing seven different emotions, including Happiness, Sadness, Neutral, Anger, Disgust, Fear and Surprise. Each emotion category contains 80 samples, each consisting of 3D coordinates provided by 25 joints.

\textbf{GSM8K}~\citep{cobbe2021training} comprises 8.5K high-quality middle-school mathematics problems, including 7.5K training examples and 1K test examples. Each problem requires 2--8 reasoning steps and can be solved using basic arithmetic operations, including addition, subtraction, multiplication, and division.

\textbf{MATH}~\citep{hendrycksmath2021} comprises 12,500 challenging mathematics competition problems spanning a broad range of domains, including algebra, geometry, and calculus. Each problem is accompanied by a detailed solution process.

\subsection{Performance Metrics}
\label{appendix:D}
To evaluate different module weight loading strategies, training schemes, and the GET framework, we follow the evaluation protocol associated with each affective model. Specifically, for AffectGPT~\citep{lian2025affectgpt}, we use the \emph{hit rate} to evaluate performance on its downstream affective tasks. For Emotion-LLaMA, performance is evaluated using the F1 score, whereas for EAI-LLM, we use accuracy (ACC). For each model, we first evaluate its corresponding downstream tasks using the designated metric and then compute the arithmetic mean across these tasks as the overall performance (\emph{Mean}). The \emph{Mean} is computed as:
\begin{equation}
\mathrm{Mean} = \frac{1}{n}\sum_{d=1}^{n} s_d,
\end{equation}
where $s_d$ denotes the performance score (e.g., HIT, WAF) on the $d$-th dataset, and $\emph{Mean}$ represents the arithmetic average of the performance across the $n$ datasets.

For AffectGPT, the \emph{hit rate} is computed as:
\begin{equation}
\mathrm{hit} = \frac{1}{N}\sum_{i=1}^{N}\mathbb{I}\left[ G_{w_k}(y_i) \in G_{w_k}(\hat{Y}_i) \right],
\end{equation}
where $\mathbb{I}[\cdot]$ denotes an indicator function. Since emotion description tasks may produce free-form outputs $\hat{Y}_i$ containing multiple emotion labels, including finer-grained categories beyond the basic emotion set, the hit rate measures whether the predicted output contains the ground-truth basic emotion label $y_i$.

\bibliographystyle{plainnat}
\bibliography{references}

@article{shao2025reasons,
title={Who Reasons in the Large Language Models?},
author={Jie Shao and Jianxin Wu},
journal={The Thirty-ninth Annual Conference on Neural Information Processing Systems},
year={2025},

}

@article{cheng2024emotion,
  title={Emotion-llama: Multimodal emotion recognition and reasoning with instruction tuning},
  author={Cheng, Zebang and Cheng, Zhi-Qi and He, Jun-Yan and Wang, Kai and Lin, Yuxiang and Lian, Zheng and Peng, Xiaojiang and Hauptmann, Alexander},
  journal={Advances in Neural Information Processing Systems},
  volume={37},
  pages={110805--110853},
  year={2024}
}

@article{lian2025affectgpt,
  title={Affectgpt: A new dataset, model, and benchmark for emotion understanding with multimodal large language models},
  author={Lian, Zheng and Chen, Haoyu and Chen, Lan and Sun, Haiyang and Sun, Licai and Ren, Yong and Cheng, Zebang and Liu, Bin and Liu, Rui and Peng, Xiaojiang and others},
  journal={arXiv preprint arXiv:2501.16566},
  year={2025}
}

@inproceedings{huang2025emodetective,
  title={EmoDETective: Detecting, Exploring, and Thinking Emotional Cause in Videos},
  author={Huang, Xuandong and Zhou, Yuzhe and Li, Jiashu and Lu, Shiqian and Wang, Shangfei},
  booktitle={Proceedings of the 33rd ACM International Conference on Multimedia},
  pages={5735--5744},
  year={2025}
}

@article{hu2021loralowrankadaptationlarge,
      title={Lora: Low-rank adaptation of large language models.},
  author={Hu, Edward J and Shen, Yelong and Wallis, Phillip and Allen-Zhu, Zeyuan and Li, Yuanzhi and Wang, Shean and Wang, Lu and Chen, Weizhu and others},
  journal={ICLR},
  year={2022}
}

@article{song2025emotion,
  title={Emotion-o1: Adaptive long reasoning for emotion understanding in llms},
  author={Song, Changhao and Zhang, Yazhou and Gao, Hui and Huang, Kaiyun and Zhang, Peng},
  journal={arXiv preprint arXiv:2505.22548},
  year={2025}
}

@inproceedings{10.1145/3581783.3612836,
author = {Lian, Zheng and Sun, Haiyang and Sun, Licai and Chen, Kang and Xu, Mingyu and Wang, Kexin and Xu, Ke and He, Yu and Li, Ying and Zhao, Jinming and Liu, Ye and Liu, Bin and Yi, Jiangyan and Wang, Meng and Cambria, Erik and Zhao, Guoying and Schuller, Bj\"{o}rn W. and Tao, Jianhua},
title = {MER 2023: Multi-label Learning, Modality Robustness, and Semi-Supervised Learning},
year = {2023},
isbn = {9798400701085},
publisher = {Association for Computing Machinery},
booktitle = {Proceedings of the 31st ACM International Conference on Multimedia},
pages = {9610–9614},
numpages = {5},
location = {Ottawa ON, Canada},
}

@inproceedings{10.1145/3689092.3689959,
author = {Lian, Zheng and Sun, Haiyang and Sun, Licai and Wen, Zhuofan and Zhang, Siyuan and Chen, Shun and Gu, Hao and Zhao, Jinming and Ma, Ziyang and Chen, Xie and Yi, Jiangyan and Liu, Rui and Xu, Kele and Liu, Bin and Cambria, Erik and Zhao, Guoying and Schuller, Bj\"{o}rn W. and Tao, Jianhua},
title = {MER 2024: Semi-Supervised Learning, Noise Robustness, and Open-Vocabulary Multimodal Emotion Recognition},
year = {2024},
publisher = {Association for Computing Machinery},
booktitle = {Proceedings of the 2nd International Workshop on Multimodal and Responsible Affective Computing},
pages = {41–48},
numpages = {8},
location = {Melbourne VIC, Australia},
}

@article{2008IEMOCAP,
  title={IEMOCAP: interactive emotional dyadic motion capture database},
  author={ Busso, Carlos  and  Bulut, Murtaza  and  Lee, Chi Chun  and  Kazemzadeh, Abe  and  Mower, Emily  and  Kim, Samuel  and  Chang, Jeannette N.  and  Lee, Sungbok  and  Narayanan, Shrikanth S. },
  journal={Language Resources and Evaluation},
  volume={42},
  number={4},
  pages={335-359},
  year={2008},
}

@inproceedings{poria-etal-2019-meld,
    title = "{MELD}: A Multimodal Multi-Party Dataset for Emotion Recognition in Conversations",
    author = "Poria, Soujanya  and
      Hazarika, Devamanyu  and
      Majumder, Navonil  and
      Naik, Gautam  and
      Cambria, Erik  and
      Mihalcea, Rada",
    booktitle = "Proceedings of the 57th Annual Meeting of the Association for Computational Linguistics",
    month = jul,
    year = "2019",
    publisher = "Association for Computational Linguistics",
    pages = "527--536",
}

@inproceedings{zadeh-etal-2017-tensor,
    title = "Tensor Fusion Network for Multimodal Sentiment Analysis",
    author = "Zadeh, Amir  and
      Chen, Minghai  and
      Poria, Soujanya  and
      Cambria, Erik  and
      Morency, Louis-Philippe",
    booktitle = "Proceedings of the 2017 Conference on Empirical Methods in Natural Language Processing",
    month = sep,
    year = "2017",
    publisher = "Association for Computational Linguistics",
    pages = "1103--1114",
}

@inproceedings{bagher-zadeh-etal-2018-multimodal,
    title = "Multimodal Language Analysis in the Wild: {CMU}-{MOSEI} Dataset and Interpretable Dynamic Fusion Graph",
    author = "Bagher Zadeh, AmirAli  and
      Liang, Paul Pu  and
      Poria, Soujanya  and
      Cambria, Erik  and
      Morency, Louis-Philippe",
    
    booktitle = "Proceedings of the 56th Annual Meeting of the Association for Computational Linguistics",
    month = jul,
    year = "2018",
    publisher = "Association for Computational Linguistics",
    pages = "2236--2246",
}

@inproceedings{yu-etal-2020-ch,
    title = "{CH}-{SIMS}: A {C}hinese Multimodal Sentiment Analysis Dataset with Fine-grained Annotation of Modality",
    author = "Yu, Wenmeng  and
      Xu, Hua  and
      Meng, Fanyang  and
      Zhu, Yilin  and
      Ma, Yixiao  and
      Wu, Jiele  and
      Zou, Jiyun  and
      Yang, Kaicheng",
    booktitle = "Proceedings of the 58th Annual Meeting of the Association for Computational Linguistics",
    month = jul,
    year = "2020",
    publisher = "Association for Computational Linguistics",
    pages = "3718--3727",
}

@misc{liu2022make,
    title={Make Acoustic and Visual Cues Matter: CH-SIMS v2.0 Dataset and AV-Mixup Consistent Module}, 
    author={Yihe Liu and Ziqi Yuan and Huisheng Mao and Zhiyun Liang and Wanqiuyue Yang and Yuanzhe Qiu and Tie Cheng and Xiaoteng Li and Hua Xu and Kai Gao},
    year={2022},
    eprint={2209.02604},
    archivePrefix={arXiv},
    primaryClass={cs.MM}
  }

@inproceedings{lu2025understanding,
  title={Understanding emotional body expressions via large language models},
  author={Lu, Haifeng and Chen, Jiuyi and Liang, Feng and Tan, Mingkui and Zeng, Runhao and Hu, Xiping},
  booktitle={Proceedings of the AAAI Conference on Artificial Intelligence},
  volume={39},
  number={2},
  pages={1447--1455},
  year={2025}
}

@article{yang2025echoes,
  title={Echoes of empathy: a symbiotic IoT-based emotion feedback framework for psychological interventions via large language model},
  author={Yang, Minqiang and Yang, Zhichao and Ning, Zhaolong and Shen, Hao and Mao, Chengsheng and Ma, Changsheng and Hu, Bin},
  journal={IEEE Internet of Things Journal},
  volume={12},
  number={20},
  pages={41297--41308},
  year={2025},
  publisher={IEEE}
}

@article{sofroniew2026emotion,
  title={Emotion concepts and their function in a large language model},
  author={Sofroniew, Nicholas and Kauvar, Isaac and Saunders, William and Chen, Runjin and Henighan, Tom and Hydrie, Sasha and Citro, Craig and Pearce, Adam and Tarng, Julius and Gurnee, Wes and others},
  journal={arXiv preprint arXiv:2604.07729},
  year={2026}
}

@article{wang2025llms,
  title={Do llms" feel"? emotion circuits discovery and control},
  author={Wang, Chenxi and Zhang, Yixuan and Yu, Ruiji and Zheng, Yufei and Gao, Lang and Song, Zirui and Xu, Zixiang and Xia, Gus and Zhang, Huishuai and Zhao, Dongyan and others},
  journal={arXiv preprint arXiv:2510.11328},
  year={2025}
}

@article{liu2026escrag,
  title={ESCRAG-R1: Retrieval-Augmented Reinforcement Learning for Emotional Support Conversation},
  author={Liu, Weichu and Hu, Yuxuan and Sun, Yirong and Mao, Ningning and Zhang, Ziyun and Chen, Jian and Xu, Mingyang and Zhong, Qishan and Li, Chengming},
  journal={arXiv preprint arXiv:2608.21925},
  year={2026}
}

@article{chen2026towards,
  title={Towards stable cross-domain depression recognition under missing modalities},
  author={Chen, Jiuyi and Tan, Mingkui and Lu, Haifeng and Xu, Qiuna and Wang, Zhihua and Zeng, Runhao and Hu, Xiping},
  journal={Pattern Recognition},
  pages={113367},
  year={2026},
  publisher={Elsevier}
}

@article{Fourati_Pelachaud_2018,   title={Perception of Emotions and Body Movement in the Emilya Database},  volume={9},  DOI={10.1109/taffc.2016.2591039},  number={1},  journal={IEEE Transactions on Affective Computing},  author={Fourati, Nesrine and Pelachaud, Catherine},  year={2018},  month={Jan},  pages={90–101},  language={en-US}  }

@inproceedings{sapinski2019multimodal,
  title={Multimodal database of emotional speech, video and gestures},
  author={Sapi{\'n}ski, Tomasz and Kami{\'n}ska, Dorota and Pelikant, Adam and Ozcinar, Cagri and Avots, Egils and Anbarjafari, Gholamreza},
  booktitle={Pattern Recognition and Information Forensics: ICPR 2018 International Workshops, CVAUI, IWCF, and MIPPSNA, Beijing, China, August 20-24, 2018, Revised Selected Papers 24},
  pages={153--163},
  year={2019},
  organization={Springer}
}

@article{Zhang_Yu_Zhang_Du_Zhan_Chen_Jiang_Guo_Zhao_Wang_etal._2020,   title={Kinematic dataset of actors expressing emotions},  DOI={10.1038/s41597-020-00635-7},  journal={Scientific Data},  author={Zhang, Mingming and Yu, Lu and Zhang, Keye and Du, Bixuan and Zhan, Bin and Chen, Shaohua and Jiang, Xiuhao and Guo, Shuai and Zhao, Jiafeng and Wang, Yang and Wang, Bin and Liu, Shenglan and Luo, Wenbo},  year={2020},  month={Sep},  language={en-US}  }

@article{meng2024pissa,
  title={Pissa: Principal singular values and singular vectors adaptation of large language models},
  author={Meng, Fanxu and Wang, Zhaohui and Zhang, Muhan},
  journal={Advances in Neural Information Processing Systems},
  volume={37},
  pages={121038--121072},
  year={2024}
}

@article{anderson2001lesions,
  title={Lesions of the human amygdala impair enhanced perception of emotionally salient events},
  author={Anderson, Adam K and Phelps, Elizabeth A},
  journal={Nature},
  volume={411},
  number={6835},
  pages={305--309},
  year={2001},
  publisher={Nature Publishing Group UK London}
}

@article{janak2015circuits,
  title={From circuits to behaviour in the amygdala},
  author={Janak, Patricia H and Tye, Kay M},
  journal={Nature},
  volume={517},
  number={7534},
  pages={284--292},
  year={2015},
  publisher={Nature Publishing Group UK London}
}

@article{dettmers2023qlora,
  title={Qlora: Efficient finetuning of quantized llms},
  author={Dettmers, Tim and Pagnoni, Artidoro and Holtzman, Ari and Zettlemoyer, Luke},
  journal={Advances in neural information processing systems},
  volume={36},
  pages={10088--10115},
  year={2023}
}

@article{cobbe2021training,
  title={Training verifiers to solve math word problems},
  author={Cobbe, Karl and Kosaraju, Vineet and Bavarian, Mohammad and Chen, Mark and Jun, Heewoo and Kaiser, Lukasz and Plappert, Matthias and Tworek, Jerry and Hilton, Jacob and Nakano, Reiichiro and others},
  journal={arXiv preprint arXiv:2110.14168},
  year={2021}
}

@article{hendrycksmath2021,
  title={Measuring Mathematical Problem Solving With the MATH Dataset},
  author={Dan Hendrycks and Collin Burns and Saurav Kadavath and Akul Arora and Steven Basart and Eric Tang and Dawn Song and Jacob Steinhardt},
  journal={NeurIPS},
  year={2021}
}

@article{zhang2023adalora,
  title={Adalora: Adaptive budget allocation for parameter-efficient fine-tuning},
  author={Zhang, Qingru and Chen, Minshuo and Bukharin, Alexander and Karampatziakis, Nikos and He, Pengcheng and Cheng, Yu and Chen, Weizhu and Zhao, Tuo},
  journal={arXiv preprint arXiv:2303.10512},
  year={2023}
}

@inproceedings{zhou2025lora,
  title={Lora-drop: Efficient lora parameter pruning based on output evaluation},
  author={Zhou, Hongyun and Lu, Xiangyu and Xu, Wang and Zhu, Conghui and Zhao, Tiejun and Yang, Muyun},
  booktitle={Proceedings of the 31st international conference on computational linguistics},
  pages={5530--5543},
  year={2025}
}

@inproceedings{wei2025flexora,
  title={Flexora: Flexible low-rank adaptation for large language models},
  author={Wei, Chenxing and Shu, Yao and He, Ying Tiffany and Yu, Fei},
  booktitle={Proceedings of the 63rd Annual Meeting of the Association for Computational Linguistics},
  pages={14643--14682},
  year={2025}
}

@inproceedings{ouyang2025k,
  title={K-lora: Unlocking training-free fusion of any subject and style loras},
  author={Ouyang, Ziheng and Li, Zhen and Hou, Qibin},
  booktitle={2025 IEEE/CVF Conference on Computer Vision and Pattern Recognition (CVPR)},
  pages={13041--13050},
  year={2025},
  organization={IEEE}
}

@article{huang2026oneemo,
  title={OneEmo: A Unified Multimodal Reasoning Model for Emotion Perception, Understanding, and Interaction},
  author={Huang, Jiahao and Lian, Zheng and Zhang, Jingyi and Chen, Zhide and Peng, Xiaojiang and Wang, Shaonan},
  journal={arXiv preprint arXiv:2608.06013},
  year={2026}
}

@inproceedings{poria2017context,
  title={Context-dependent sentiment analysis in user-generated videos},
  author={Poria, Soujanya and Cambria, Erik and Hazarika, Devamanyu and Majumder, Navonil and Zadeh, Amir and Morency, Louis-Philippe},
  booktitle={Proceedings of the 55th annual meeting of the association for computational linguistics (volume 1: Long papers)},
  pages={873--883},
  year={2017}
}

% \appendix
% \input{sections/appendix}

\end{document}